\documentclass[runningheads]{llncs}
\usepackage{graphicx}

\usepackage{tikz}
\usepackage{comment} 
\usepackage{amsmath,amssymb}
\usepackage{color}

\usepackage[width=122mm,left=12mm,paperwidth=146mm,height=193mm,top=12mm,paperheight=217mm]{geometry}
\usepackage[final]{pdfpages}

\usepackage[font=small,labelfont=bf,margin=\parindent,tableposition=top]{caption}

\usepackage{xspace}

\usepackage[caption=false]{subfig}

\usepackage{colortbl}
\definecolor{Gray}{gray}{0.8}
\usepackage{multirow}
\usepackage{rotating}
\usepackage{hhline}

\usepackage{xcolor}

\usepackage{url}

\newcommand{\highlight}[1]{{\textcolor{black}{{#1}}}}
\newcommand{\highlightNUMB}[1]{{\textcolor{black}{{#1}}}}
\newcommand{\highlightFIGTAB}[1]{\mbox{\textcolor{black}{{#1}}}}

\newcommand{\highlightSEC}[1]{\mbox{\textcolor{black}{{#1}}}}

\newcommand{\grabTitle}{GRAB: A Dataset of Whole-Body\\Human Grasping of Objects}

\newcommand{\video}{{{video}}\xspace}

\newcommand{\etal}{et al.\xspace}
\newcommand{\ie}{i.e.\xspace}
\newcommand{\eg}{e.g.\xspace}

\newcommand{\mano}{\mbox{MANO}\xspace}

\newcommand{\grab}{GRAB\xspace}

\newcommand{\supmat}{{Sup. Mat.}\xspace}

\newcommand{\graspIt}{\mbox{GraspIt}\xspace}

\newcommand{\shapenet}{\mbox{ShapeNet}\xspace}

\newcommand{\grabnet}{\mbox{GrabNet}\xspace}
\newcommand{\coarsenet}{\mbox{CoarseNet}\xspace}
\newcommand{\refnet}{\mbox{RefineNet}\xspace}
\newcommand{\grabnetlatentD}{{16}\xspace}

\newcommand{\mocap}{\mbox{MoCap}\xspace}

\newcommand{\sota}{{state-of-the-art}\xspace}

\newcommand{\twoD}{{2D}\xspace}
\newcommand{\threeD}{\xspace{3D}\xspace}

\newcommand{\rgb}{RGB\xspace}
\newcommand{\rgbD}{\mbox{RGB-D}\xspace}

\newcommand{\vicon}{{Vicon}\xspace}
\newcommand{\shogun}{\xspace{Sh\={o}gun}\xspace}

\newcommand{\shogunPost}{{\shogun-Post}\xspace}

\newcommand{\markersALLL}{{\highlightNUMB{$99$}}\xspace}
\newcommand{\markersBODY}{{\highlightNUMB{$49$}}\xspace}
\newcommand{\markersFACE}{{\highlightNUMB{$14$}}\xspace}

\newcommand{\markersHANDs}{{\highlightNUMB{$36$}}\xspace}
\newcommand{\markersBODYsize}{{\highlightNUMB{$4.5$ mm}}\xspace}
\newcommand{\markersHANDsize}{{\highlightNUMB{$1.5$ mm}}\xspace}

\newcommand{\markersOBJJsize}{\markersHANDsize}

\newcommand{\dataSubjects}{{\highlightNUMB{$10$}}\xspace}
\newcommand{\dataSubjectsHALF}{{\highlightNUMB{$5$}}\xspace}

\newcommand{\smplX}{\mbox{SMPL-X}\xspace}
\newcommand{\moshpp}{\mbox{MoSh++}\xspace}
\newcommand{\mosh}{\moshpp}

\newcommand{\citeMOSH}{\mbox{\cite{AMASS:2019}}\xspace}
\newcommand{\printerAMD}{{{Stratasys Fortus $360$mc}}\xspace}

\newcommand{\mesh}{M}

\newcommand{\obj}{o}

\newcommand{\body}{b}
\newcommand{\face}{f}
\newcommand{\hand}{h}

\newcommand{\trans}{{\gamma}}

\newcommand{\iring}{\mathcal{R}}
\newcommand{\imesh}{\mathcal{\mesh}}
\newcommand{\contact}{\mathcal{C}}

\DeclareSymbolFont{matha}{OML}{txmi}{m}{it}
\DeclareMathSymbol{\varv}{\mathord}{matha}{118}

\newcommand{\markerDistBody}{d_b}
\newcommand{\markerDistHand}{d_h}
\newcommand{\markerDistFace}{d_f}
\newcommand{\markerDistBodyVAL}{9.5}

\newcommand{\betasNumb}{\highlightNUMB{10}}
\newcommand{\ncomps}{\highlight{60}}
\newcommand{\ncompsPerHand}{\xspace\highlight{30}\xspace}

\newcommand{\contactDBobjects}{{\highlightNUMB{$51$}}\xspace}
 
\newcommand{\websiteURL}{\mbox{\url{https://grab.is.tue.mpg.de}}}

\begin{document}
\pagestyle{headings}
\mainmatter

\title{\grabTitle}
\titlerunning{\grabTitle}
\author{
Omid Taheri			\and	
Nima Ghorbani		\and
Michael J. Black		\and	
Dimitrios Tzionas
}

\authorrunning{O. Taheri \etal}

\institute{
\scriptsize
Max Planck Institute for Intelligent Systems, T{\"u}bingen, Germany
\email{\{otaheri,nghorbani,black,dtzionas\}@tuebingen.mpg.de}
}

\maketitle

\begin{abstract}
Training computers to understand, model, and synthesize human grasping requires a rich dataset containing complex \threeD object shapes, detailed contact information, hand pose and shape, and the \threeD body motion over time.
While ``grasping'' is commonly thought of as a single hand stably lifting an object, we capture the motion of the entire body and adopt the generalized notion of ``\mbox{whole-body} grasps''. 
Thus, we collect a new dataset, called {\em GRAB} (GRasping Actions with Bodies), of whole-body grasps, containing full \threeD shape and pose sequences
of \dataSubjects subjects interacting with \contactDBobjects everyday objects of varying shape and size. 
Given \mocap markers, we fit the full \threeD body shape and pose, including the articulated face and hands, as well as the \threeD object pose. 
This gives detailed \threeD meshes over time, from which we compute contact between the body and object. 
This is a unique dataset, that goes well beyond existing ones for modeling and understanding how humans grasp and manipulate objects, how their full body is involved, and how interaction varies with the task.
We illustrate the practical value of GRAB with an example application; we train \grabnet, a conditional generative network, to predict \threeD hand grasps for unseen \threeD object shapes.
The dataset and code are available for research purposes at \websiteURL.
\end{abstract}

\section{Introduction}	\label{sec:intro}

\begin{figure}[t]
    \centerline{
	\includegraphics[trim=000mm 000mm 000mm 000mm, clip=true, width=1.00 \linewidth]{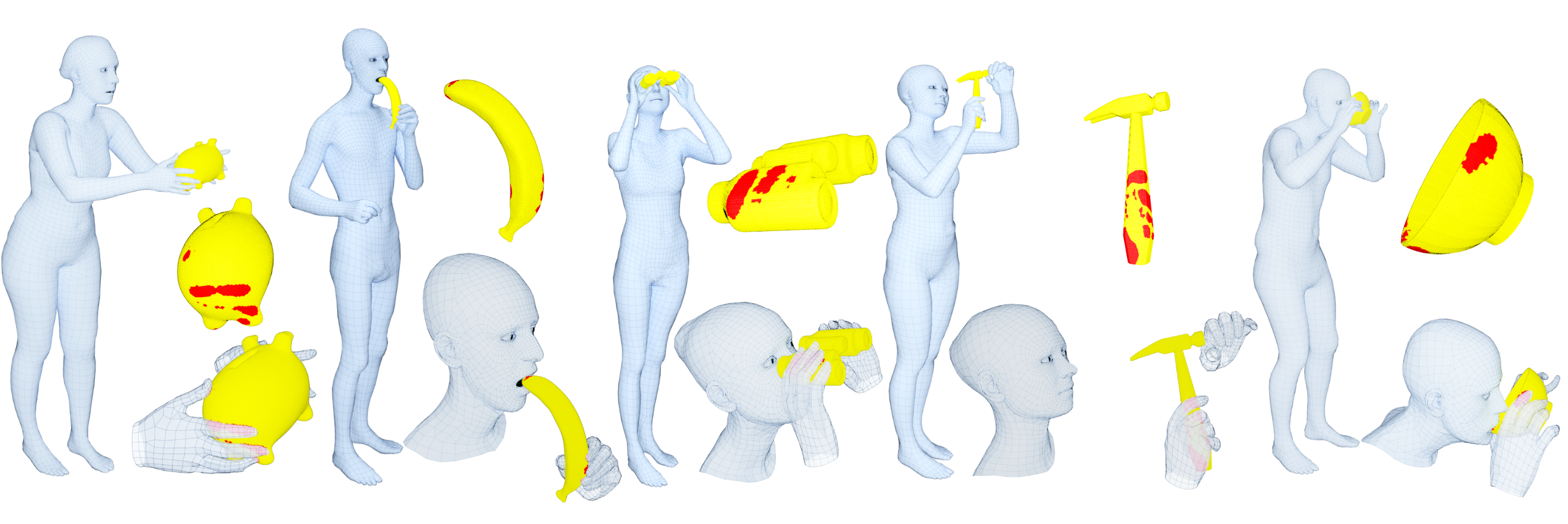}
	}
	\caption{
						Example ``whole-body grasps'' from the GRAB dataset. 
						A ``grasp'' is usually thought of as a single hand interacting with an object. 
						Using objects, however, may involve  more than just a single hand.
						From left to right: 
						(i)		passing a piggy bank, 
						(ii)		eating a banana, 
						(iii)	looking through binoculars, 
						(iv)		using a hammer, 
						(v)		drinking from a bowl. 
						Contact between the object and the body is shown in red on the object; here contact areas are spatially extended to aid visualization. 
						See the \video on our website for a wide range of sequences with various objects and intents. 
	}
	\label{fig:teaser}
\end{figure}

A key goal of computer vision is to estimate human-object interactions from video to help understand human behavior.
Doing so requires a strong model of such interactions and learning this model requires data.
However, capturing such data is not simple.
Grasping involves both gross and subtle motions, as humans involve their whole body and dexterous finger motion to manipulate objects. 
Therefore, objects contact multiple body parts and not just the hands.  
This is difficult to capture with images because the regions of contact are occluded.
Pressure sensors or other physical instrumentation, however, are also not a full solution as they can impair natural human-object interaction and do not capture full-body motion.
Consequently, there are no existing datasets of complex human-object interaction that contain full-body motion, \threeD body shape, and detailed body-object contact.
To fill this gap, we capture a novel dataset of full-body \threeD humans dynamically interacting with \threeD objects as illustrated in Fig. \ref{fig:teaser}.
By accurately tracking \threeD body and object shape, we reason about contact resulting in a dataset with detail and richness beyond existing grasping datasets.

Most previous work focuses on prehensile ``grasps'' \cite{napier1956prehensile}; i.e.~a single human hand stably lifting or using an object.
The hands, however, are only part of the story. 
For example, as infants, our earliest grasps involve bringing objects to the mouth \cite{ruff1984infants}.
Consider the example of  drinking from a bowl in Fig. \ref{fig:teaser} (right). 
To do so, we must pose our body so that we can reach it, we orient our head to see it, we move our arm and hand to stably lift it, and then we bring it to our mouth, making contact with the lips, and finally we tilt the head to drink.
As this and other examples in the figure illustrate, human grasping and using of everyday objects involves the \emph{whole body}. 
Such interactions are fundamentally \emph{three-dimensional}, and contact occurs between objects and multiple body parts.

{\bf Dataset.}
Such whole-body grasping \cite{hsiao2006wholeBodyGrasp} has received much less attention \cite{asfour2015wholeTaxonomy,Laptvev_CVPR_2019_forces} 
than single hand-object grasping \cite{dillmann2005grasp,cutkosky1989grasp,Feix_GRASP_2016,kamakura1980grasp,napier1956prehensile}.
To model such  grasping  we need a dataset of humans interacting with varied objects, capturing the full \threeD surface of both the body and objects.
To solve this problem we adapt recent motion capture techniques, to construct a new rich dataset called {\bf \grab} for ``\emph{GRasping Actions with Bodies.}''
Specifically, we adapt \moshpp \citeMOSH in two ways.
First, \moshpp estimates the \threeD shape and motion of the body and hands from \mocap markers; here we extend this to include facial motion.  
For increased accuracy we first capture a \threeD scan of each subject and fit the  \smplX body model \cite{smplifyPP} to it.  Then \moshpp is used to recover the pose of the body, hands and face.
Note that the face is important because it is involved in many interactions; see in Fig. \ref{fig:teaser} (second from left) how the mouth opens to eat a banana. 
Second, we also accurately capture the motion of \threeD objects as they are manipulated by the subjects.
To this end, we use small hemi-spherical markers on the objects and show that these do not impact grasping behavior.
As a result, we obtain detailed \threeD meshes for both the object and the human (with a full body, articulated fingers and face) moving over time while in interaction, as shown in Fig. \ref{fig:teaser}. 
Using these meshes we then infer the body-object contact (red regions in Fig. \ref{fig:teaser}).  
Unlike Brahmbhatt \etal \cite{contactDB_2019} this gives both the contact and the full body/hand pose over time. 
Interaction is dynamic, including \mbox{in-hand} manipulation and \mbox{re-grasping}. 
\grab captures \dataSubjects different people (\dataSubjectsHALF male and \dataSubjectsHALF female) interacting with \contactDBobjects objects from \cite{contactDB_2019}.
Interaction takes place in $4$ different contexts: lifting, handing over, passing from one hand to the other, and using, depending on the affordances and functionality of the object.

{\bf Applications.}
\grab supports multiple uses of interest to the community.
First, we show how \grab can be used to gain insights into hand-object contact in everyday scenarios.
Second, there is a significant interest in training models to grasp \threeD objects \cite{bohg2014graspSurvey,SahbaniEB12}.
Thus, we use \grab to train a pair of neural networks (coarse prediction followed by refinement) to generate plausible grasps for unseen \threeD objects. 
Given a randomly posed \threeD object, we predict plausible hand parameters ({\color{black} $6$ DoF} wrist pose and {\color{black} full} finger articulation) appropriate for grasping the object.
To encode arbitrary \threeD object shapes, we employ the recent basis point set (BPS) representation \cite{BPS19}, whose fixed size is appropriate for neural networks. 
Then, by conditioning on a new \threeD object shape, we sample from the learned latent space, and generate hand grasps for this. 
We quantitatively and qualitatively evaluate the resulting grasps and show that they look natural. 

In summary, this work makes the following contributions:
(1) we introduce a unique dataset capturing real ``whole-body grasps'' of \threeD objects, including \mbox{full-body} human motion, object motion, \mbox{in-hand} manipulation and \mbox{re-grasps};
(2) to capture this, we adapt \moshpp to solve for the body, face and hands of \smplX ~to obtain detailed moving \threeD meshes;
(3) using these meshes and tracked \threeD objects we compute plausible contact on the object and the human and provide an analysis of observed patterns; 
(4) we show the value of our dataset for machine learning, by training a novel conditional neural network to generate \threeD hand grasps for unseen \threeD objects.
The dataset, models, and code are available for research purposes at \websiteURL.

\section{Related Work}	\label{sec:related}

\textbf{Hand Grasps:}
Hands are crucial for grasping and manipulating objects. 
For this reason, many studies focus on understanding grasps and defining taxonomies \mbox{\cite{dillmann2005grasp,cutkosky1989grasp,Feix_GRASP_2016,kamakura1980grasp,napier1956prehensile,pressureprofileWEB}}, by exploring 
the object shape and purpose of grasps \cite{cutkosky1989grasp}, 
the contact areas on the hand captured by sinking objects in ink \cite{kamakura1980grasp}, 
the pose and contact areas \cite{dillmann2005grasp} captured with an integrated \mbox{data-glove} \cite{cybergloveWEB} and \mbox{tactile-glove} \cite{pressureprofileWEB}, or 
the number of fingers in contact with the object and thumb position \cite{Feix_GRASP_2016}. 
A key element for these studies is capturing accurate hand poses, relative \mbox{hand-object} configurations and contact areas. 

\textbf{Whole-Body Grasps:}
Often people use more than a single hand to interact with objects. 
However, there are not many works in the literature on this topic \mbox{\cite{asfour2015wholeTaxonomy,hsiao2006wholeBodyGrasp}}. 
Borras et al. \cite{asfour2015wholeTaxonomy} use \mocap data \cite{kitMocap2015} of people interacting with a scene with \mbox{multi-contact}, and present a body pose taxonomy for such whole-body grasps. 
Hsiao et al. \cite{hsiao2006wholeBodyGrasp} focus on imitation learning with
a database of whole-body grasp demonstrations with a human teleoperating a simulated robot. 
Although these works go in the right direction, they use unrealistic humanoid models and 
simple objects \cite{asfour2015wholeTaxonomy,hsiao2006wholeBodyGrasp} or synthetic ones \cite{hsiao2006wholeBodyGrasp}. 
Instead, we use the \smplX model \cite{smplifyPP} to capture ``whole-body'', face and dexterous \mbox{in-hand} interactions. 

\textbf{Capturing Interactions with \mocap:}
\mocap is often used to capture, synthesize or evaluate humans interacting with scenes.
Lee \etal \cite{lee2006motionPatches} capture a \threeD body skeleton interacting with a \threeD scene and show how to synthesize new motions in new scenes.
Wang \etal \cite{wang2019geometric} capture a \threeD body skeleton interacting with large geometric objects. 
Han \etal \cite{oculus2018handMarkers} present a method for automatic labeling of hand markers, to speed up hand tracking for VR. 
Le \etal \cite{henze_2018_comfort} capture a hand interacting with a phone to study the ``comfortable areas'', while 
Feit \etal \cite{feit2016howWeType} capture two hands interacting with a keyboard to study typing patterns. 
Other works \cite{kry2006interaction,pollard2005physically} focus on graphics applications. 
Kry \etal \cite{kry2006interaction} capture a hand interacting with a \threeD shape primitive, instrumented with a force sensor. 
Pollard \etal \cite{pollard2005physically} capture the motion of a hand to learn a controller for physically based grasping. 
Mandery \etal \cite{kitMocap2015} sit between the above works, capturing humans interacting with both big and handheld objects, but without articulated faces and fingers. 
None of the previous work captures full \threeD bodies, hands and faces together with \threeD object manipulation and contact. 

\textbf{Capturing Contact:} 
Capturing human-object contact is hard, because the human and object heavily occlude each other. 
One approach is instrumentation with touch and pressure sensors, but this might bias natural grasps. 
Pham \etal \cite{pham2018pami} predefine contact points on objects to place force transducers.
Recent advances in tactile sensors allow accurate recognition of tactile patterns and handheld objects \cite{torralba_2019_natureGlove}. 
Some approaches \cite{dillmann2005grasp} use a data glove \cite{cybergloveWEB} with an embedded tactile glove \cite{pressureprofileWEB,tecksanGripWEB}, but this combination is complicated and the two modalities can be hard to synchronize. 
A microscopic-domain tactile sensor \cite{gelsightWEB} is introduced in \cite{gelsight2011siggraph}, but is not easy to use on human hands.
Mascaro \etal \cite{mascaro2001photoplethysmograph} attach a minimally invasive camera to detect changes in the coloration of fingernails. 
Brahmbhatt \etal \cite{contactDB_2019} use a thermal camera to directly observe the ``thermal print'' of a hand on the grasped object. 
However, for this they only capture static grasps that last long enough for heat transfer. 
Consequently, even recent datasets that capture realistic \mbox{hand-object} \cite{FirstPersonAction_CVPR2018,hampali2019ho} or \mbox{body-scene} \mbox{\cite{PROX:2019,savva2016pigraphs}} interaction avoid directly measuring contact. 

\textbf{\threeD Interaction Models:} 
Learning a model of \mbox{human-object} interactions is useful for graphics and robotics to help avatars \mbox{\cite{elkoura2003handrix,garciahern2020physicsbased,starke2019neuralStateMachine}} or 
robots \cite{handa2019dexpilot} interact with their surroundings, and for vision \cite{Corona_2020_CVPR,Hamer_ObjectPrior,iMapper2018} to help reconstruct interactions from ambiguous data.  
However, this is a \mbox{chicken-and-egg} problem; to capture or synthesize data to learn a model, one needs such a model in the first place. 
For this reason, the community has long used \mbox{hand-crafted} approaches that exploit contact and physics, for 
\mbox{body-scene} 	\mbox{\cite{Brubaker2009,hasler_2009_unsyncMovCam,PROX:2019,rosenhahn2008,Yamamoto2000}}, 
\mbox{body-object} 	\mbox{\cite{shape2pose2014,Black_TrackPople,Laptvev_CVPR_2019_forces}}, or 
\mbox{hand-object} 	\mbox{\cite{hasson_2019_obman,Oikonomidis_1hand_object,Rogez:ICCV:2015,JavierHandsInAction,srinath_eccv2016_handObject,Tsoli:2018:ECCV,Tzionas:IJCV:2016,wang2013handObject,liu2012synthesisHandonly,zhang2019interactionfusion}} scenarios. 
These approaches compute contact approximately. 
This approximation may be rough when humans are modeled as \threeD skeletons \cite{shape2pose2014,Laptvev_CVPR_2019_forces} or shape primitives \cite{Brubaker2009,Black_TrackPople,srinath_eccv2016_handObject}. 
However, it gets relatively accurate when using \threeD meshes 
that are generic \cite{Rogez:ICCV:2015,Tsoli:2018:ECCV}, 
personalized \cite{hasler_2009_unsyncMovCam,rosenhahn2008,Tzionas:IJCV:2016,wang2013handObject}, 
or based on \threeD statistical models \mbox{\cite{PROX:2019,hasson_2019_obman}}. 

To collect training data, several works \cite{Rogez:ICCV:2015,JavierHandsInAction} use synthetic Poser \cite{poserWEB} hand models, manually articulated to grasp \threeD shape primitives. 
Contact points and forces are also annotated \cite{Rogez:ICCV:2015} through proximity and \mbox{inter-penetration} of \threeD meshes. 
In contrast, Hasson \etal \cite{hasson_2019_obman} use the robotics method \graspIt \cite{Miller2004} to automatically generate \threeD \mano \cite{romero2017embodied} grasps for \shapenet~\cite{ShapeNet2015} objects and render synthetic images of the hand-object interaction. 
However, \graspIt optimizes for \mbox{hand-crafted} grasp metrics that do not necessarily reflect the distribution of human grasps (see \mbox{Sup.~Mat.}~Sec.~\mbox{C.2} of \cite{hasson_2019_obman}, and \cite{GraspDatabase2009}). 
Alternatively, Garcia-Hernando \etal \cite{FirstPersonAction_CVPR2018} use magnetic sensors to reconstruct a \threeD hand skeleton and rigid object poses; they capture $6$ subjects interacting with $4$ objects. 
This data is used for \threeD hand/object pose estimation \cite{kokic2019learning,Tekin_2019_CVPR} or motion generation \cite{garciahern2020physicsbased}, but suffers from noisy poses and severe \mbox{inter-penetrations} (see Sec.~\mbox{5.2} of \cite{hasson_2019_obman}). 

For bodies, Kim \etal \cite{shape2pose2014} use synthetic data to learn to detect contact points on a \threeD object, and then fit  an interacting \threeD body skeleton to them. 
Savva \etal \cite{savva2016pigraphs} use \rgbD to capture \threeD body skeletons of $5$ subjects interacting in $30$ \threeD scenes, 
to learn to synthesize interactions \cite{savva2016pigraphs}, affordance detection \cite{scenegrok2014savva}, or to reconstruct interaction from videos \cite{iMapper2018}. 
Mandery \etal \cite{kitMocap2015} use optical \mocap to capture  $43$ subjects interacting with $41$ tracked objects, both large and small. 
This is similar to our effort but they do not capture fingers or \threeD body shape, so cannot reason about contact.
Corona \etal \cite{corona2020context} use this dataset to learn \mbox{context-aware} body motion prediction. 
Starke \etal \cite{starke2019neuralStateMachine} use Xsens IMU sensors \cite{xsensWEB} to capture the main body of a subject interacting with large objects, and learn to synthesize avatar motion in virtual worlds. 
Hassan \etal \cite{PROX:2019} use \rgbD and \threeD scene constraints to capture $20$ humans as \smplX \cite{smplifyPP} meshes interacting with $12$ static \threeD scenes, but do not capture object manipulation. 
Zhang \etal \cite{zhang2020generating} use this data to learn to generate \threeD \mbox{scene-aware} humans. 

We see that only parts of our problem have been studied. 
We draw inspiration from prior work, in particular \mbox{\cite{contactDB_2019,PROX:2019,hsiao2006wholeBodyGrasp,kitMocap2015}}. 
We go beyond these by introducing a new dataset of real ``\mbox{{whole-body}}'' grasps, as described in the next section. 

\section{Dataset}	\label{sec:data}

To manipulate an object, the human needs to approach its \threeD surface, and bring their skin to come in \emph{physical contact} to apply forces. 
Realistically capturing such \mbox{human-object} interactions, especially with ``\mbox{whole-body} grasps'', is a challenging problem. 
First,	the object may occlude the body and \mbox{vice-versa}, resulting in \emph{ambiguous} observations. 
Second,	for physical interactions it is crucial to reconstruct an accurate and detailed \threeD \emph{surface} for both the human and the object. 
Additionally, the capture has to work across multiple scales (body, fingers and face) and for objects of varying complexity.
We address these challenges with a unique combination of \mbox{\sota} solutions that we adapt to the problem. 

There is a fundamental trade-off with current technology; one has to choose between 
(a) accurate motion with instrumentation and without natural \rgb images, or 
(b) less accurate motion but with \rgb images. 
Here we take the former approach; for an extensive discussion we refer the reader to \supmat

\newcommand{\sizMarkersetsObj}{0.370}
\begin{figure}[t]
	\centering
	\includegraphics[trim=000mm 000mm 000mm 000mm, clip=false, 	height=\sizMarkersetsObj \linewidth]{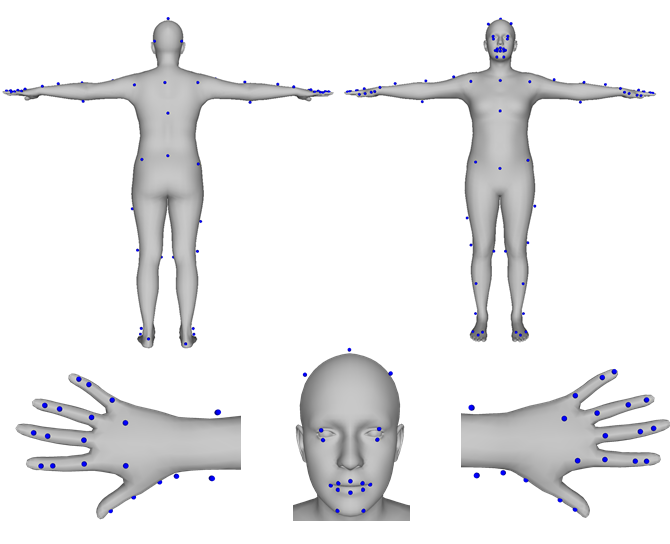}										\hspace{-0.5em}
	\includegraphics[trim=000mm 000mm 000mm 000mm, clip=false, 	height=\sizMarkersetsObj \linewidth]{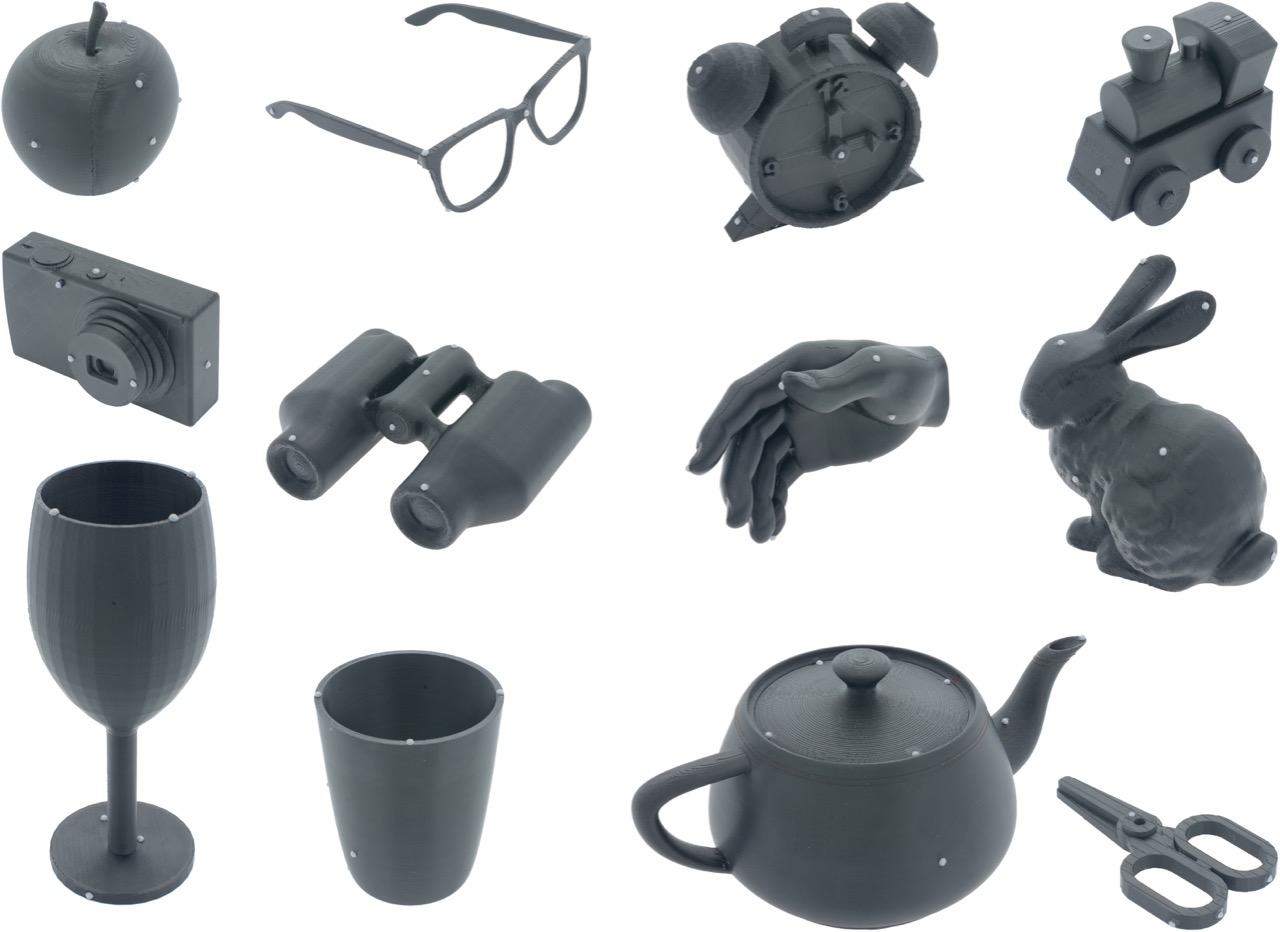}
	\caption{
				\mocap markers used to capture humans and objects.
				\textbf{Left:} We attach \markersALLL reflective markers per subject;
				\markersBODY for the body, \markersFACE for the face and \markersHANDs for the fingers.  
				We use spherical \markersBODYsize radius markers for the body and \mbox{hemi-spherical} \markersHANDsize radius ones for the hands and face.
				\textbf{Right:} Example \threeD printed objects from \cite{contactDB_2019}.  We glue \markersOBJJsize radius \mbox{hemi-spherical} markers (the gray dots) on the objects.
				These makers are small enough to be unobtrusive.
				The 6 objects on the right are mostly used by one or more hands, while the 6 on the left involve ``whole-body grasps''.  
	}
	\label{fig:dataset_markers_humans_objects}
\end{figure}

\newcommand{\sizDataSeq}{0.48}
\newcommand{\horDataSeq}{+0.8em}
\begin{figure}[t]
	\centering
	\includegraphics[trim=000mm 000mm 000mm 000mm, clip=false, width=\sizDataSeq \linewidth]{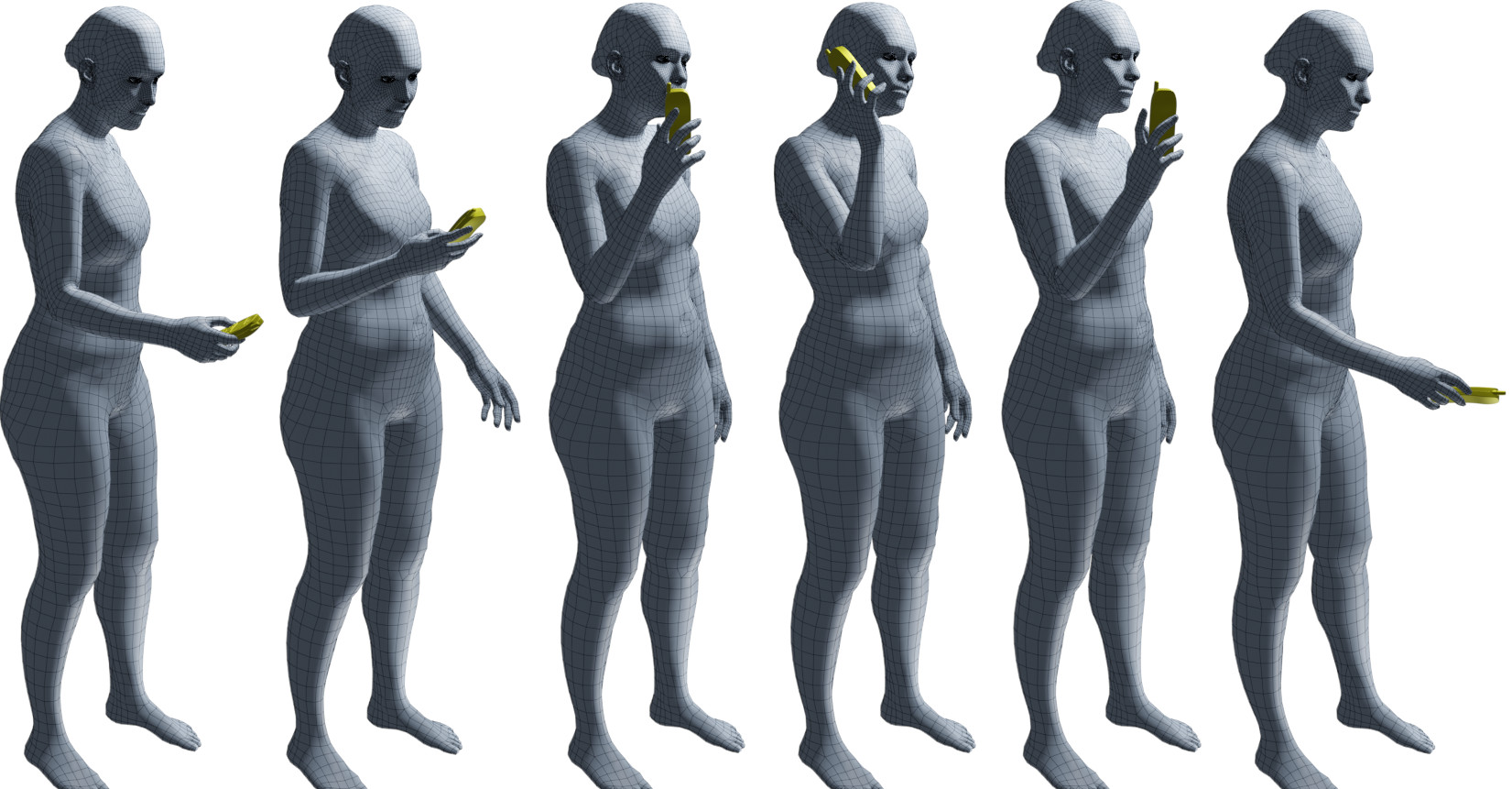}					\hspace{\horDataSeq}
	\includegraphics[trim=000mm 000mm 000mm 000mm, clip=false, width=\sizDataSeq \linewidth]{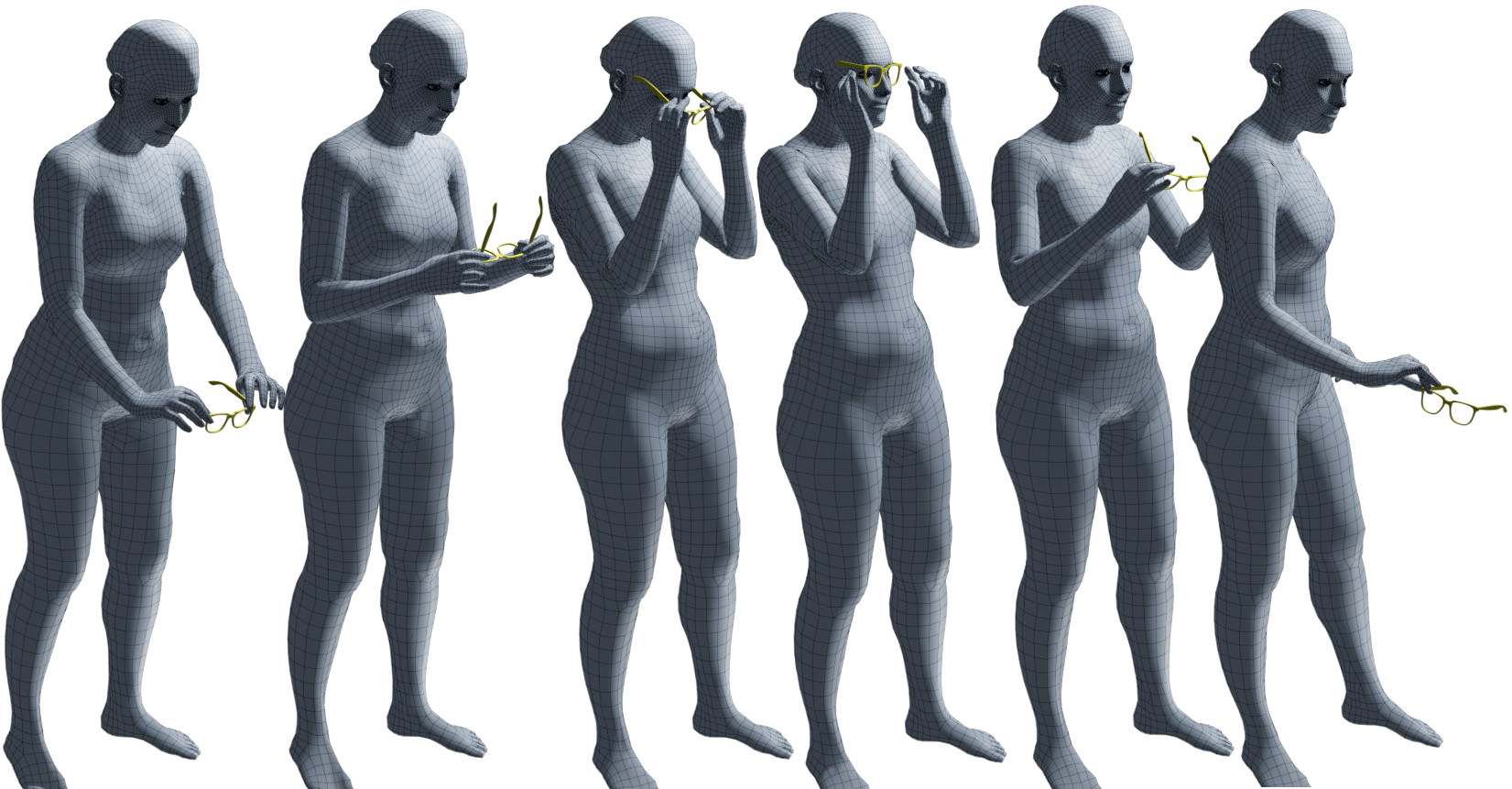}			\\	
	\includegraphics[trim=000mm 000mm 000mm 000mm, clip=false, width=\sizDataSeq \linewidth]{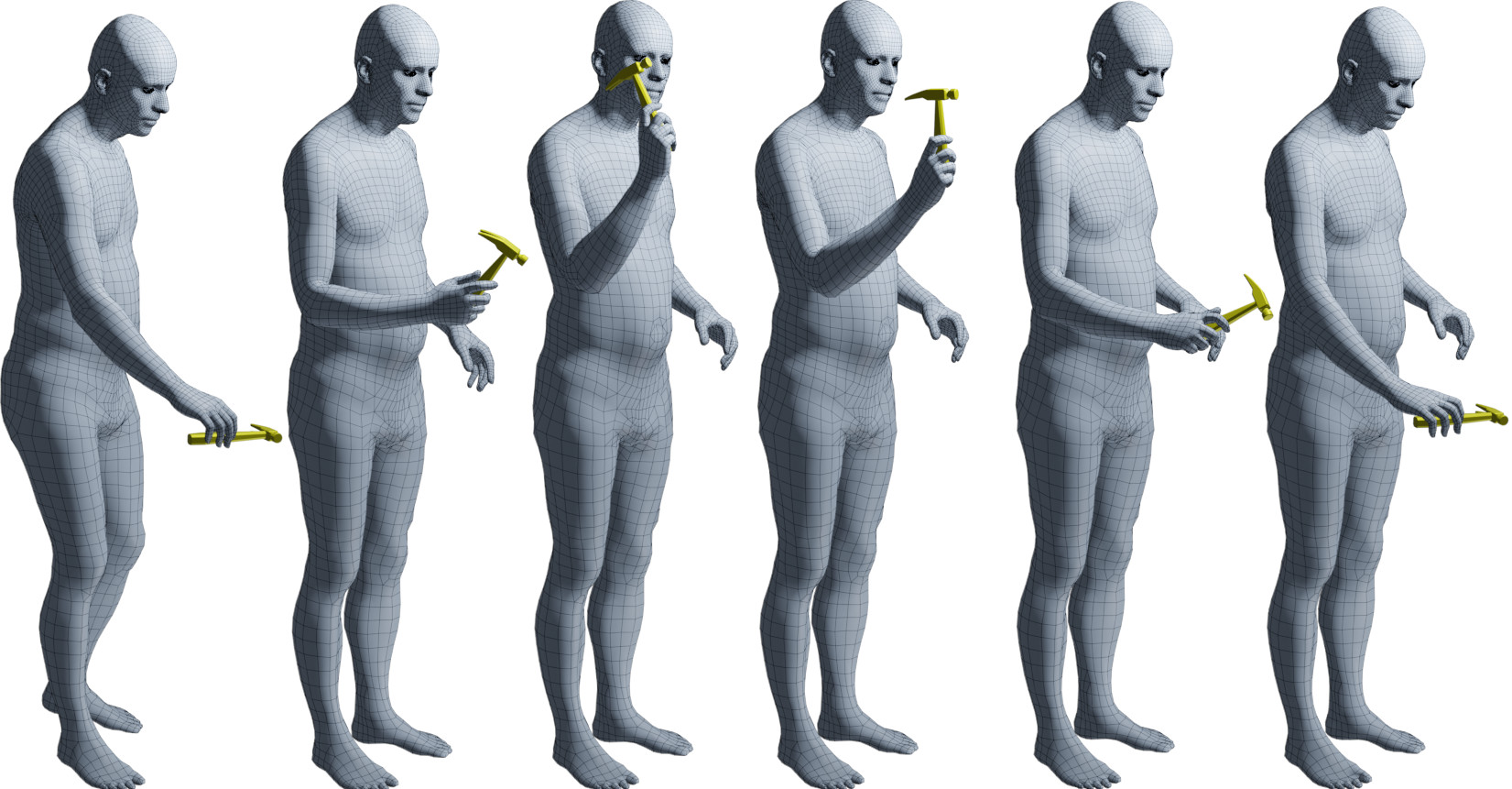}					\hspace{\horDataSeq}
	\includegraphics[trim=000mm 000mm 000mm 000mm, clip=false, width=\sizDataSeq \linewidth]{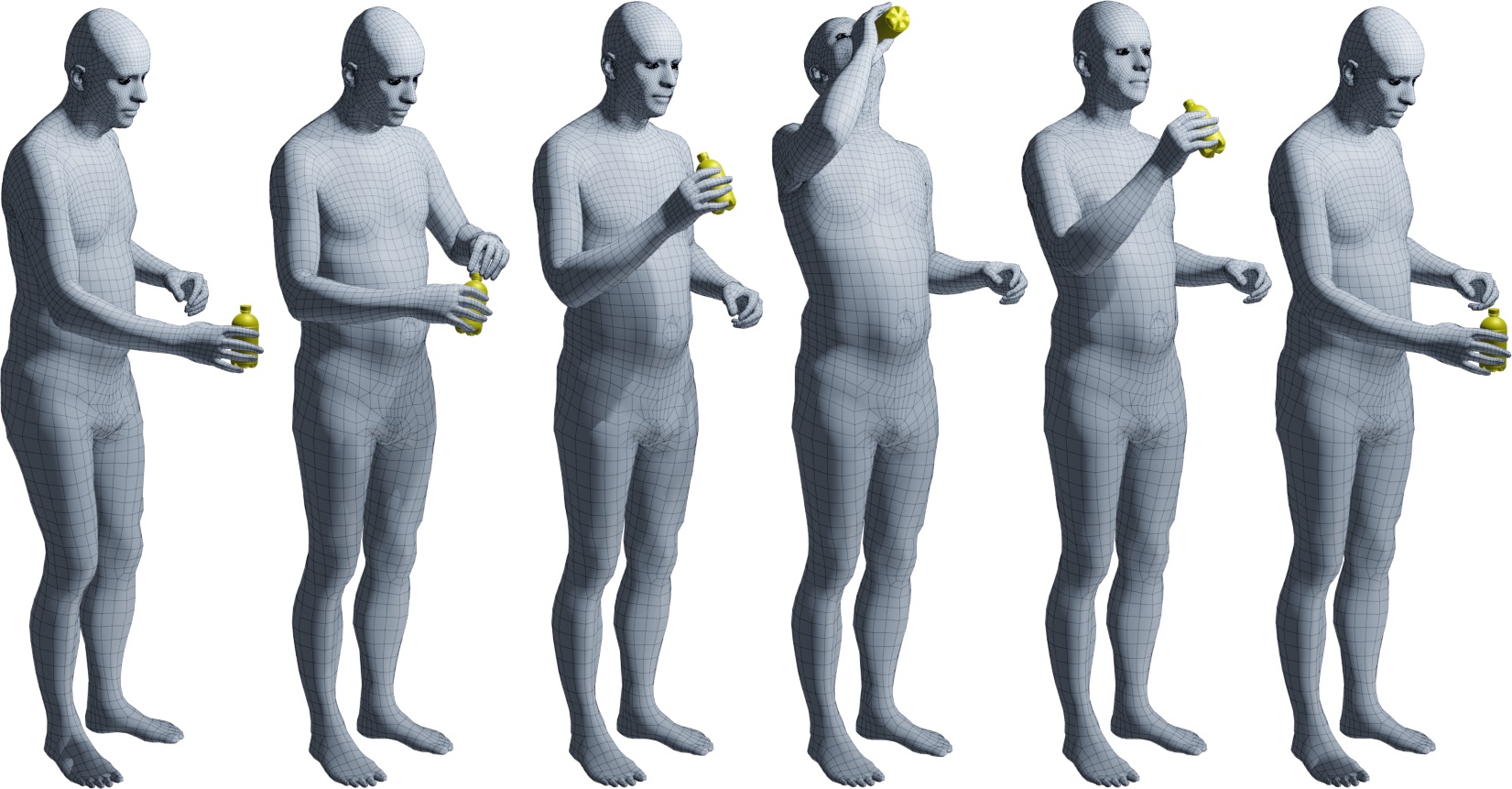}				
	\caption{
						We capture humans interacting with objects over time and reconstruct sequences of \threeD meshes for both, as described in \highlightSEC{Sec. \ref{sec:dataset_mocap}} and \highlightSEC{Sec. \ref{sec:dataset_mocap2surface}}. 
						Note the realistic and plausible placement of objects in the hands, and the ``\mbox{whole-body}'' involvement. 
						The \video on our website shows more examples. 
	}
	\label{fig:dataset_sequences}
\end{figure}

\subsection{Motion Capture (MoCap)}					\label{sec:dataset_mocap}				We use a \vicon system with $54$ infrared ``Vantage $16$'' \cite{viconVantageWEB} cameras that capture $16$ MP at $120$ fps. 
The large number of cameras minimizes occlusions and the high frame rate captures temporal details of contact. 
The high resolution allows to capture even small (\markersHANDsize~\highlight{radius}) hemi-spherical markers. 
This minimizes their influence on finger and face motion and does not alter how people grasp objects.
Details of the marker setup are shown in Fig.~\ref{fig:dataset_markers_humans_objects}.
Even with many cameras, motion capture of the body, face, and hands, together with objects, is uncommon because it is so challenging. 
\mocap markers become occluded, labels are swapped, and ghost makers appear.
\mocap cleaning was done by four trained technicians using Vicon's \shogunPost software.

\textbf{Capturing Human \mocap:}
To capture human motion, we use the marker set of \highlightFIGTAB{Fig. \ref{fig:dataset_markers_humans_objects} (left)}. 
The body markers are attached on a tight body suit with \mbox{velcro-based} mounting at a distance of roughly $\markerDistBody = \markerDistBodyVAL$ mm from the body surface. 
The hand and face markers are attached directly on the skin with special removable glue, therefore the distance to it is roughly $\markerDistHand = \markerDistFace \approx 0$ mm. 
Importantly, no hand glove is used and hand markers are placed only on the dorsal side, leaving the palmar side completely uninstrumented, for natural interactions. 

\textbf{Capturing Objects:}
To reconstruct interactions accurately, it is important to know the precise \threeD object surface geometry. 
We therefore use the CAD object models of \cite{contactDB_2019}, and \threeD print them with a \printerAMD ~\cite{stratasysFortus360WEB} printer; see \highlightFIGTAB{Fig.~\ref{fig:dataset_markers_humans_objects} (right)}. 
Each object $o$ is then represented by a known \threeD mesh with vertices $V_{\obj}$.
To capture object motion, we attach \markersOBJJsize radius \mbox{hemi-spherical} markers directly on the object surface with strong glue.
We use at least $8$ markers per object, empirically distributing them on the object so that at least $3$ of them are always observed. 
The size and placement of the markers makes them unobtrusive.
In \supmat we show empirical evidence that makers have minimal influence on grasping. 

\subsection{From MoCap Markers to \threeD Surfaces}	\label{sec:dataset_mocap2surface}		\textbf{Human Model:} 
We model the human with the \smplX \cite{smplifyPP} \threeD body model.
\smplX jointly models the body with an articulated face and fingers; this expressive body model is critical to capture physical interactions. 
More formally, \smplX is a differentiable function $\mesh_\body(\beta, \theta, \psi, \trans)$ that is parameterized by body shape $\beta$, pose $\theta$, facial expression $\psi$ and translation $\trans$. 
The output is a \threeD mesh $\mesh_\body = (V_{\body}, F_{\body})$ with $N_{\body}=10475$ vertices $V_{\body} \in \mathbb{R}^{(N_{\body} \times 3)}$ and triangles $F_{\body}$. 
The shape parameters $\beta \in \mathbb{R}^{\betasNumb}$ are coefficients in a learned \mbox{low-dimensional} linear shape space. 
This lets \smplX represent different subject identities with the same mesh topology. 
The \threeD joints, $J(\beta)$, of a kinematic skeleton are regressed from the body shape defined by $\beta$.
The skeleton has $55$ joints in total; $22$ for the body, $15$ joints per hand for finger articulation, and $3$ for the neck and eyes. 
Corrective blend shapes are added to the body shape, and then the posed body is defined by linear blend skinning with this underlying skeleton. 
The overall pose parameters $\theta=(\theta_\body, \theta_\face, \theta_\hand)$ are comprised of $\theta_\body \in \mathbb{R}^{66}$ and $\theta_\face \in \mathbb{R}^{9}$ parameters in axis-angle representation for the main body and face joints correspondingly, 
with $3$ degrees of freedom (DoF) per joint, and $\theta_\hand \in \mathbb{R}^\ncomps$ parameters in a lower-dimensional pose space for both hands, \ie \ncompsPerHand DoF per hand following \cite{hasson_2019_obman}. 
For more details, please see \cite{smplifyPP}. 

\textbf{Model-Marker Correspondences:} 
For the human body we define, a priori, the rough marker placement on the body as shown in \highlightFIGTAB{Fig.~\ref{fig:dataset_markers_humans_objects}} \highlightFIGTAB{(left)}. 
Exact marker locations on individual subjects are then computed automatically using \mosh \citeMOSH.
In contrast to the body, the objects have different shapes and mesh topologies.
Markers are placed according to the object shape, affordances and expected occlusions during interaction; see \highlightFIGTAB{Fig.~\ref{fig:dataset_markers_humans_objects}} \highlightFIGTAB{(right)}. 
Thus, we annotate \mbox{object-specific} \mbox{vertex-marker} correspondences, and do this once per object.

\textbf{Human and Object Tracking:} 
To ensure accurate human shape, we capture a \threeD scan of each subject and fit \smplX to it following \cite{romero2017embodied}. 
We fit these personalized \smplX models to our cleaned \threeD marker observations using \mosh \citeMOSH.
Specifically we optimize over pose, \mbox{$\theta$}, expressions, $\psi$, and translation, $\trans$, while keeping the known shape, $\beta$, fixed.
The weights of \mosh for the finger and face data terms are tuned on a synthetic dataset, as described in \supmat 
An analysis of \mosh fitting accuracy is also provided in \supmat

Objects are simpler because they are rigid and we know their \threeD shape.
Given three or more detected markers, we solve for the rigid object pose $\theta_\obj \in \mathbb{R}^{6}$. 
Here we track the human and object separately and on a per-frame basis.
\highlightFIGTAB{Figure \ref{fig:dataset_sequences}} shows that our approach captures realistic interactions and reconstructs detailed \threeD meshes for both the human and the object, over time. 
The \video on our website shows a wide range of reconstructed sequences. 

\subsection{Contact Annotation}						\label{sec:dataset_contactAnnotation}	Since contact cannot be directly observed, we estimate it using \threeD proximity between the \threeD human and object meshes. 
In theory, they come in contact when the distance between them is zero. 
In practice, however, we relax this and define contact when the distance, $d$, is below a tolerance, threshold $d \leq \epsilon_{contact}$.
This helps address:
(1)	measurement and fitting errors, 
(2)	limited mesh resolution, 
(3)	the fact that human soft tissue deforms when grasping an object, while the \smplX model cannot model this. 

\newcommand{\sizDataContactComp}{0.285}
\begin{figure}[t]
	\centering	
	\begin{minipage}{1.0\textwidth}
	\includegraphics[trim=000mm 000mm 070mm 000mm, clip=true, height=\sizDataContactComp \linewidth]{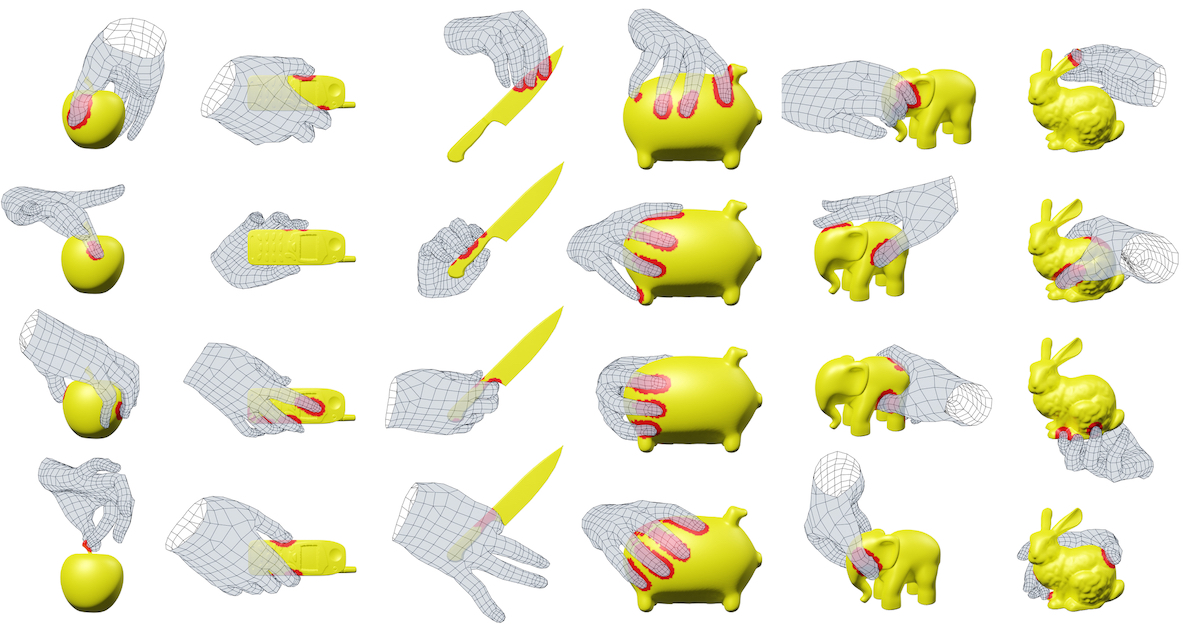}
	\includegraphics[trim=003mm 004mm 003mm 015mm, clip=true, height=\sizDataContactComp \linewidth]{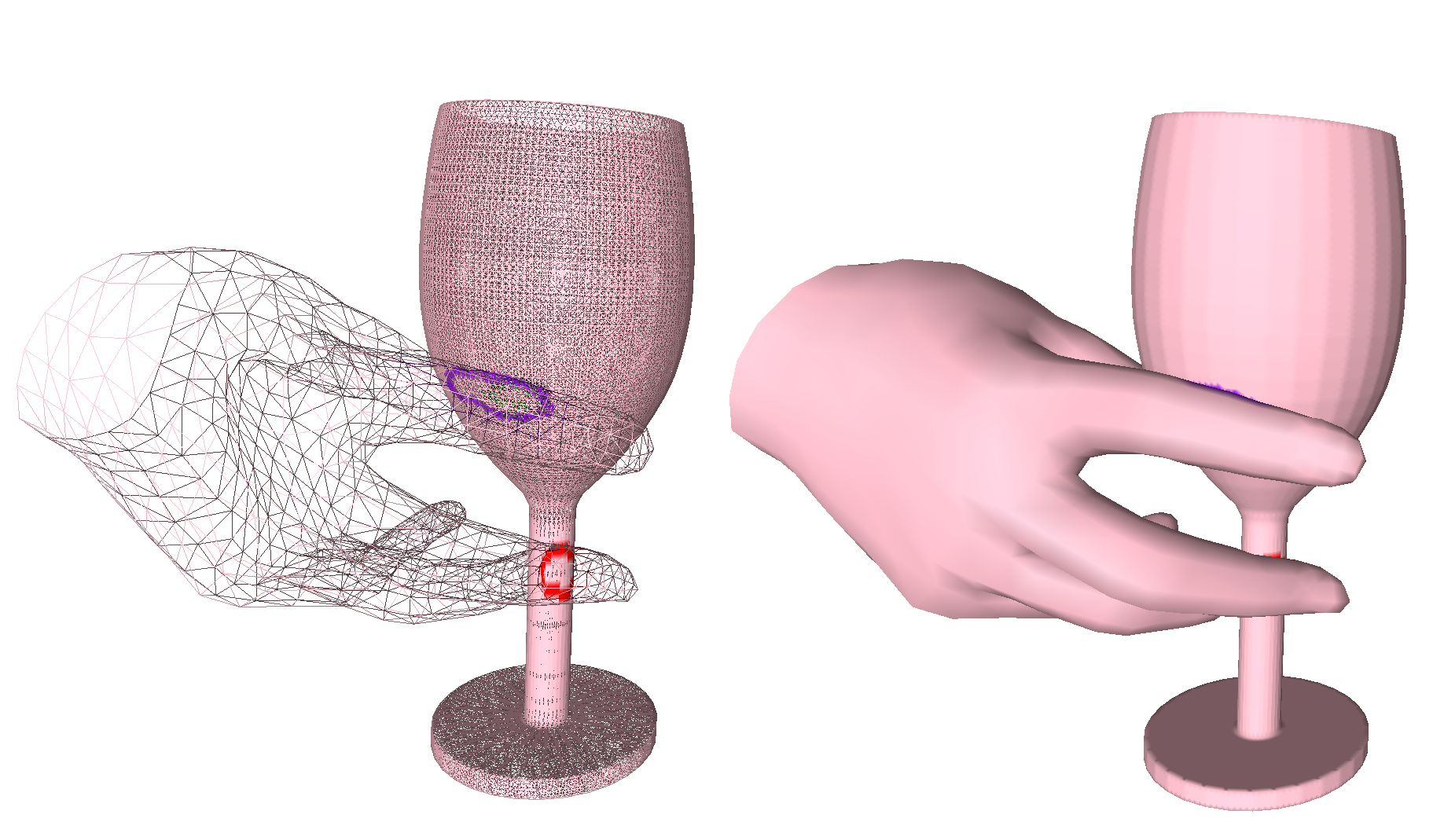}
	\end{minipage}
	\caption{
							\textbf{Left:}
							Accurate tracking lets us compute realistic contact areas (red) for each frame (\highlightSEC{Sec. \ref{sec:dataset_contactAnnotation}}).
							For illustration, we render only the hand of \smplX and spatially extend the red contact areas for visibility.  
							\textbf{Right:}
							Detection of ``intersection ring'' triangles during contact annotation (\highlightSEC{Sec. \ref{sec:dataset_contactAnnotation}}).
	}
	\label{fig:dataset_contact_Computation_Clusters}
\end{figure}

Given these issues, accurately estimating contact is challenging.
Consider the hand grasping a wine glass in \highlightFIGTAB{Fig. \ref{fig:dataset_contact_Computation_Clusters}} ~\highlightFIGTAB{(right)}, where the color rings indicate intersections. 
Ideally, the glass should be in contact with the thumb, index and middle fingers. 
``Contact \mbox{under-shooting}'' 	results in fingers hovering close to the object surface, but not on it, like the  thumb.  
``Contact \mbox{over-shooting}'', 	results in fingers penetrating the object surface around the contact area, like the index (purple intersections) and middle finger (red intersections). 
The latter case is especially problematic for thin objects where a penetrating finger can pass through the object, intersecting it on two sides.
In this example, we want to annotate contact only with the outer surface of the object and not the inner one. 

We account for ``contact \mbox{over-shooting}'' cases with an efficient heuristic. 
We use a fast method \cite{Karras:2012:MPC,smplifyPP} to detect intersections, cluster them in connected ``intersection rings'', $\iring_{\body} \subsetneq V_{\body}$ and $\iring_{\obj} \subsetneq V_{\obj}$, and label them with the intersecting body part, 
seen as purple and red rings in \highlightFIGTAB{Fig. \ref{fig:dataset_contact_Computation_Clusters}} ~\highlightFIGTAB{(right)}. 
The ``intersection ring'',  $\iring_{\body}$, segments the body mesh, $\mesh_{\body}$, to give the ``penetrating \mbox{sub-mesh}'', $\imesh_{\body} \subsetneq \mesh_{\body}$. 
We then identify two cases: 
(1) When a body part gives only one intersection, 
we      annotate as contact points on the object, $V_{\obj}^{\contact}  \subset V_{\obj}$,  all vertices enclosed by the ring $\iring_{\obj}$.
We then annotate as contact points on the body,   $V_{\body}^{\contact} \subset V_{\body}$, all vertices that lie close to $V_{\obj}^{\contact}$ with a distance \mbox{$d_{\obj \to \body} \leq \epsilon_{contact}$}. 
(2) In case of multiple intersections, $i$, we take into account only the ring $\iring_{\body}^{i}$ corresponding to the largest intersection subset, $\imesh_{\body}^{i}$. 

For body parts that are not found in contact above, there is the possibility of  ``contact \mbox{under-shooting}''.
To address this, we compute the distance from each object vertex, $V_{\obj}$, to each non-intersecting body vertex, $V_{\body}$.
We then annotate as contact vertices, $V_{\obj}^{\contact}$ and $V_{\body}^{\contact}$, the ones with \mbox{$d_{\obj \to \body} \leq \epsilon_{contact}$}. 
We empirically find that $\epsilon_{contact} = \highlight{4.5}$ mm works well for our purposes. 

\subsection{Dataset Protocol}						\label{sec:dataset_protocol}				\mbox{Human-object} interaction depends on various factors including the human body shape, object shape and affordances, object functionality, or interaction intent, to name a few. 
We therefore capture \dataSubjects people ($5$ men and $5$ women), of various sizes and nationalities, interacting with the objects of \cite{contactDB_2019}; see example objects in \highlightFIGTAB{Fig. \ref{fig:dataset_markers_humans_objects} (right)}. 
All subjects gave informed written consent to share their data for research purposes.

For each object we capture interactions with $4$ different intents, namely 
``use'' 				and 	
``pass'' 			(to someone), borrowed from the protocol of \cite{contactDB_2019}, as well as 
``lift'' 			and	
``off-hand pass'' 	(from one hand to the other). 
Figure \ref{fig:dataset_sequences} shows some example \threeD capture sequences for the ``use'' intent. 
For each sequence we: 
(i)		we randomize initial object placement to increase motion variance, 
(ii)		we instruct the subject to follow an intent, 
(iii)	the subject starts from a \mbox{T-pose} and approaches the object, 
(iv)		performs the instructed task, and 
(v)		leaves the object and returns to a \mbox{T-pose}. 
The \video on our website shows the richness of our protocol with a wide range of captured sequences. 

\subsection{Analysis}								\label{sec:dataset_analysis}				\begin{figure}[t]
	\centering
	\includegraphics[trim=000mm 000mm 000mm 000mm, clip=false, width=1.00 \linewidth]{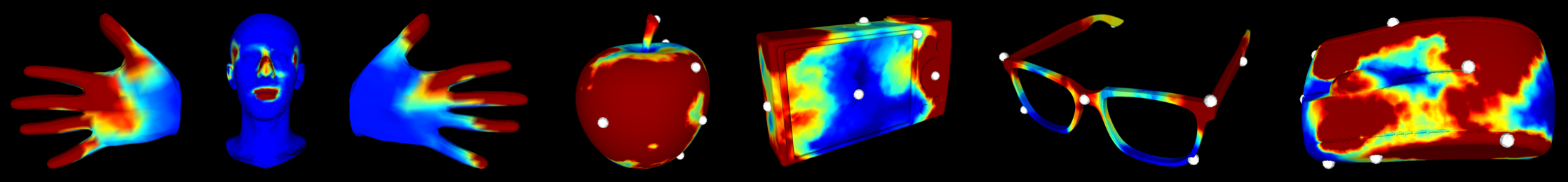}
	\caption{
					Contact ``heatmaps''. 
					\textbf{Left:}	For the body we focus on ``use'' sequences to show ``\mbox{whole-body grasps}''.
					\textbf{Right:}	For objects we include all intents.
					Object markers (light gray) are unobtrusive and can lie on ``hot'' (red) contact areas. 
	}
	\label{fig:heatmaps}
\end{figure}

\begin{table}[t]
	\begin{center}
	\caption{
						Size of the \grab dataset. 
						\grab is sufficiently
                                                large to enable
                                                training of
                                                data-driven models of
                                                grasping as shown in \highlightSEC{Sec. \ref{sec:grabnet}}. 
	}
	\label{tab:dataset_statistics}
	\scriptsize
	\setlength{\tabcolsep}{1pt}
	\begin{tabular}{cccccccc}																																																																				\hhline{-~----~-}
	\multicolumn{1}{|l|}{~Intent~}			&	\multicolumn{1}{c}{}		&	\multicolumn{1}{|c|}{~``Use''~}		&	\multicolumn{1}{c|}{~``Pass''~}		&	\multicolumn{1}{c|}{~``Lift''~}		&	\multicolumn{1}{c|}{~``Off-hand''~}	&	\multicolumn{1}{c}{}	&	\multicolumn{1}{|c|}{~Total~}		\\	\hhline{-~----~-}
	\noalign{\smallskip}																																																																						\hhline{-~----~-}
	\multicolumn{1}{|l|}{~\# Sequences~}		&	\multicolumn{1}{c}{}		&	\multicolumn{1}{|c|}{~$579$~}		&	\multicolumn{1}{c|}{~$414$~}			&	\multicolumn{1}{c|}{~$274$~}			&	\multicolumn{1}{c|}{~$67$~}			&	\multicolumn{1}{c}{}	&	\multicolumn{1}{|c|}{~$1334$~}		\\	\hhline{-~----~-}
	\multicolumn{1}{|l|}{~\# Frames~}		&	\multicolumn{1}{c}{}		&	\multicolumn{1}{|c|}{~$605.796$~}	&	\multicolumn{1}{c|}{~$335.733$~}		&	\multicolumn{1}{c|}{~$603.381$~}		&	\multicolumn{1}{c|}{~$77.549$~}		&	\multicolumn{1}{c}{}	&	\multicolumn{1}{|c|}{~$1.622.459$~} 	\\	\hhline{-~----~-}
	\end{tabular}
	\end{center}
\end{table}

\begin{figure}[t]
	\centering
	\includegraphics[trim=000mm 000mm 000mm 000mm, clip=false, width=0.70 \linewidth]{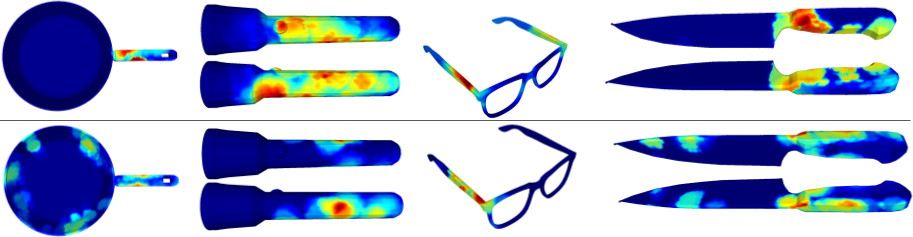}
	\caption{
						Effect of interaction intent on contact during grasping. 
						We show the ``use'' (top) and ``pass'' (bottom) intents for $4$ different objects.
	}
	\label{fig:intent_effect}
\end{figure}

\textbf{Dataset Analysis.}
The dataset contains 1334 sequences and over 1.6M frames of \mocap; 
\highlightFIGTAB{Table \ref{tab:dataset_statistics}} provides a detailed breakdown. 
Here we analyze those frames where we have detected contact between the human and the object.
We assume that the object is static on a table and can move only due to grasping.
Consequently, we consider contact frames to be those in which the object's position deviates in the vertical direction by at least $5$ mm from its initial position and in which at least $50$ body vertices are in contact with the object. 
This results in $952,514$ contact frames that we analyze below.
The exact thresholds of these contact heuristics have little influence on our analysis, see \supmat

By uniquely capturing the whole body, and not just the hand, interesting interaction patterns arise. 
By focusing on ``use'' sequences that highlight the object functionality, we observe that 
	$92\%$ of contact frames involve			the right hand, 
	$39\%$ 							the left  hand, 
	$31\%$ both hands, and 
	$8\%$ involve the head. 
For the first category the \mbox{per-finger} contact likelihood, from thumb to pinky, is $100\%$, $96\%$, $92\%$, $79\%$, $39\%$ and for the palm $24\%$. 
For more results see \supmat 

To visualize the \mbox{fine-grained} contact information, we integrate over time the binary \mbox{per-frame} contact maps, and generate ``heatmaps'' encoding the contact likelihood across the whole body surface. 
\highlightFIGTAB{Figure \ref{fig:heatmaps} (left)} shows such ``heatmaps'' for ``use'' sequences. 
``Hot'' areas (red) denote high likelihood of contact, while ``cold'' areas (blue) denote low likelihood. 
We see that both the hands and face are important for using everyday objects, highlighting the importance of capturing the whole interacting body.
For the face, the ``hot'' areas are the lips, the nose, the temporal head area, and the ears. 
For hands, the fingers are more frequently in contact than the palm, with more contact on the right hand than the left one.
The palm seems more important for \mbox{right-hand} grasps than for \mbox{left-hand} ones, possibly because all our subjects are \mbox{right-handed}. 
Contact patterns are also influenced by the size of the object and the size of the hand; see \supmat for a visualization.

\highlightFIGTAB{Figure \ref{fig:intent_effect}} shows the effect of the intent. 
Contact for ``use'' sequences complies with the functionality of the object; \eg people do not touch the knife blade or the hot area of the pan, but they do contact the on/off button of the flashlight. 
For ``pass'' sequences subjects tend to contact one side of the object irrespective of affordances, leaving the other one free to be grasped by the receiving person. 

For natural interactions it is important to have a minimally intrusive setup. 
While our \mocap markers are small and unobtrusive, as seen in \highlightFIGTAB{Figure \ref{fig:dataset_markers_humans_objects}} \highlightFIGTAB{(right)}, we ask whether subjects may be biased in their grasps by these markers. 
\highlightFIGTAB{Figure \ref{fig:heatmaps}} \highlightFIGTAB{(right)} shows contact ``heatmaps'' for some objects across all intents. 
These clearly show that markers are often located in ``hot'' areas, suggesting that subjects do not avoid grasping these locations. 
Further analysis based on K-means clustering of grasps can be found in \supmat

\section{\grabnet: Learning to Grab an Object} 	\label{sec:grabnet}

We show the value of \grab with a challenging application; 
we train on it a model that generates plausible \threeD \mano \cite{romero2017embodied} grasps for an unseen \threeD object. 
Our model, \grabnet, is comprised of two modules: coarse prediction and refinement. 
This is similar to \cite{mousavian2019graspnet}, but with several key differences.
We predict full human hand pose instead of a robotic gripper, train using captured human grasps, employ a different object representation which generalizes to new objects, and make different choices for the model and losses.
Importantly, our refinement is done by a neural network, not an optimization process. 

\begin{figure}[t]
	\centering
	\includegraphics[trim=000mm 000mm 000mm 000mm, clip=false, width=1.00 \linewidth]{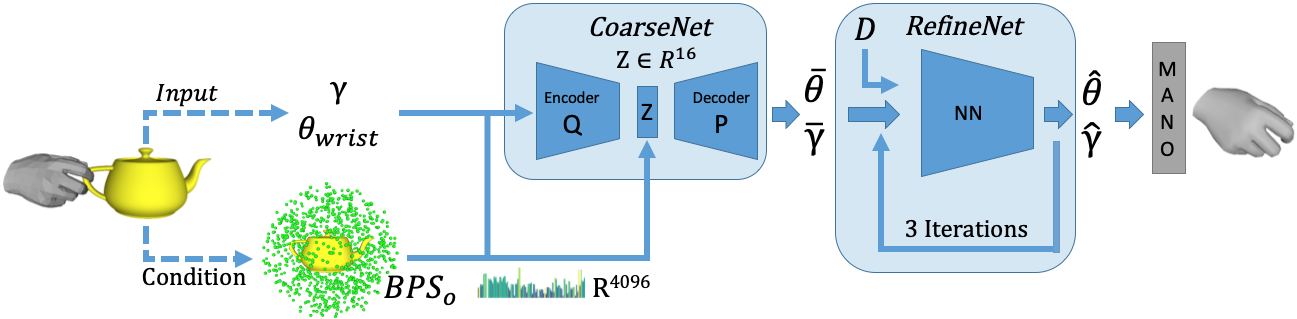}
	\caption{
						\grabnet architecture. 
						\grabnet generates \mano \cite{romero2017embodied} grasps for unseen object shapes, encoded with a BPS \cite{BPS19} representation. 
						It is comprised of two main modules. 
						First, with \coarsenet we predict an initial plausible grasp. 
						Second, we refine this with \refnet to produce better contacts with the object. 
	}
	\label{fig:grabnet_arch_simple}
\end{figure}

We first employ \coarsenet, a conditional variational autoencoder (cVAE) \cite{kingmawelling2013}, that generates an initial grasp. 
For this it learns a grasping embedding space, $Z$, conditioned on the object shape, that is encoded using the Basis Point Set (BPS) \cite{BPS19} representation as a set of distances from the basis points to the nearest object points. 
In contrast to \cite{mousavian2019graspnet}, $Z$ captures not only the $6$ DoF pre-grasp pose (gripper for \cite{mousavian2019graspnet}/wrist for \mano), but also the fully articulated human hand pose. 
\coarsenet's grasps are reasonable, but realism can improve by refining contacts based on the distances, $D$, between the hand and object meshes. 
We do this with a second network, called \refnet, that employees explicit \mano contact points learned on our \grab dataset to refine the initial pose. 
In contrast, \cite{mousavian2019graspnet} learn an evaluator of the $6$ DoF gripper pose that they differentiate to optimize the coarse pose. 
The architecture of \grabnet is shown in \highlightFIGTAB{Fig. \ref{fig:grabnet_arch_simple}}, for more details see \supmat 

\newcommand{\grabMathINNtheta}{\theta_{wrist}}
\newcommand{\grabMathINNgamma}{\gamma}
\newcommand{\grabMathCOARSEoutTheta}{\bar{\theta}}
\newcommand{\grabMathCOARSEoutGamma}{\bar{\trans}}
\newcommand{\grabMathREFINEoutTheta}{\hat{\theta}}
\newcommand{\grabMathREFINEoutGamma}{\hat{\trans}}

\textbf{Pre-processing.} 
For training, we gather all frames with \mbox{right-hand} grasps that involve some minimal contact, for details see \supmat 
We then center each training sample, \ie hand-object grasp, at the centroid of the object and compute the ${BPS}_{\obj} \in R^{\highlightNUMB{4096}}$ representation for the object, used for conditioning. 

\textbf{\coarsenet.} 
We pass the object shape ${BPS}_{\obj}$ along with initial \mano~  wrist rotation $\grabMathINNtheta$ and translation $\grabMathINNgamma$ to the 
encoder $Q(Z | \grabMathINNtheta, \grabMathINNgamma, {BPS}_{\obj})$ that produces a latent grasp code $Z \in R^{\highlightNUMB{\grabnetlatentD}}$. 
The decoder $P(\grabMathCOARSEoutTheta, \grabMathCOARSEoutGamma | Z, {BPS}_{\obj})$ maps $Z$ and ${BPS}_{\obj}$ to \mano parameters with full finger articulation $\bar{\theta}$, to generate a \threeD grasping hand. 
For the training loss, we use standard cVAE loss terms (KL divergence, weight regularizer), a data term on \mano mesh edges (L1), 
as well as a penetration and a contact loss.
For the latter, we learn candidate contact point weights from \grab, in contrast to handcrafted ones \cite{PROX:2019} or weights learned from artificial data \cite{hasson_2019_obman}.
At inference time, given an unseen object shape, ${BPS}_{\obj}$, we sample the latent space, $Z$, and decode our sample to generate a \mano grasp. 

\textbf{\refnet.} 
The grasps estimated by \coarsenet are plausible, but can be refined for improved contacts. 
For this, \refnet takes as input the initial grasp ($\grabMathCOARSEoutTheta$, $\grabMathCOARSEoutGamma$) and the distances $D$ from \mano vertices to the object mesh. 
The distances are weighted according to the vertex contact likelihood learned from \grab. 
Then, \refnet estimates refined \mano parameters ($\grabMathREFINEoutTheta$, $\grabMathREFINEoutGamma$) in $3$ iterative steps as in \cite{kanazawa_cvpr_2018}, to give the final grasp. 
To train \refnet, we generate a synthetic dataset; we sample \coarsenet grasps as ground truth and we perturb their hand pose parameters to simulate noisy input estimates. 
We use the same training losses as for \coarsenet. 

\begin{figure}[t]
	\centering
	\includegraphics[trim=000mm 000mm 000mm 000mm, clip=false, width=1.00 \linewidth]{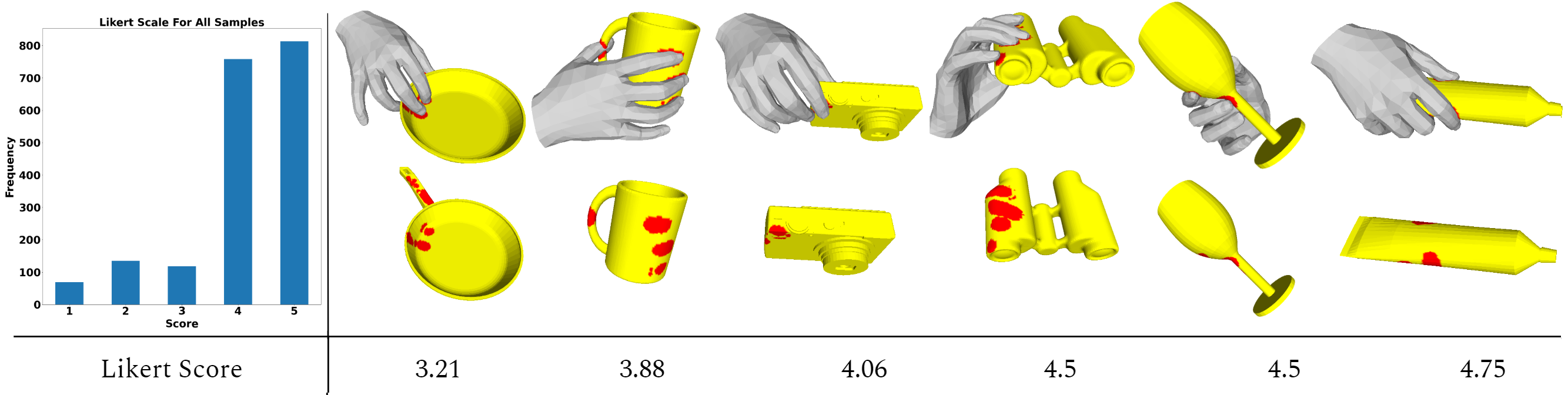}
	\caption{
								Grasps generated by \grabnet for unseen objects; grasps look natural. 
								As free \mbox{by-product} of \threeD mesh generation, we get the red contact areas. 
								For each grasp we show the average Likert score from all annotators. 
								On the left we show the average Likert score for all generated grasps (best viewed on screen).
			}
	\label{fig:grabnet_testset_gen}
\end{figure}

\textbf{\grabnet.} 
Given an unseen \threeD object, we first get an initial grasp estimate with \coarsenet, and pass this to \refnet to get the final grasp estimate.
For simplicity, the two networks are trained separately, but we expect end-to-end refinement to be beneficial, as in \cite{hasson_2019_obman}. 
\highlightFIGTAB{Figure \ref{fig:grabnet_testset_gen} (right)} shows generated examples; 
our generations look realistic, as explained later in the evaluation section. 
For more qualitative results, see the video on our website and images in \supmat

\textbf{Contact.} 
As a free \mbox{by-product} of our \threeD grasp predictions, we can compute contact between the \threeD hand and object meshes, following \highlightSEC{Sec. \ref{sec:dataset_contactAnnotation}}. 
Contacts for \grabnet are shown with red in \highlightFIGTAB{Figure \ref{fig:grabnet_testset_gen} (right)}. 
Other methods for contact prediction, like \cite{contactDB_2019}, are pure bottom-up approaches that label a vertex as in contact or not, without explicit reasoning about the hand structure. 
In contrast, we follow a top-down approach; we first generate a \threeD grasping hand, and then compute contact with explicit anthropomorphic reasoning. 

\textbf{Evaluation - \coarsenet/\refnet.}
We first quantitatively evaluate the two main components, by computing the reconstruction {vertex-to-vertex} error. 
For \coarsenet the errors are \highlight{$12.1$ mm, $14.1$ mm and $18.4$ mm} for the training, validation and test set respectively. 
For \refnet    the errors are \highlight{ $3.7$ mm,  $4.1$ mm and  $4.4$ mm}. 
The results show that the components, that are trained separately, work reasonably well before plugging them together. 

\textbf{Evaluation - \grabnet.}
To evaluate \grabnet generated grasps, we perform a user study through AMT \cite{amtWEB}. 
We take 6 test objects from the dataset and, for each object, we generate $20$ grasps, mix them with $20$ ground-truth grasps, and show them with a rotating \threeD viewpoint to subjects. 
Then we ask participants how they agree with the statement ``Humans can grasp this object as the video shows'' on a $5$-level Likert scale ($5$ is ``strongly agree'' and $1$ is ``strongly disagree''). 
To filter out the noisy subjects, namely the ones who do not understand the task or give random answers, 
we use catch trials that show implausible grasps.  
We remove subjects who rate these catch trials as realistic; see \supmat for details. 
Table \ref{tab:grabnet_evaluation} (left) shows the user scores for both ground-truth and generated grasps.

\newcommand{\sizZZZ}{+0.5em}
\begin{table}[t]
	\begin{center}
	\caption{
					\grabnet evaluation for 6 test objects.
					The ``AMT'' column shows user study results; grasp quality is rated from 1 (worst) to 5 (best). 
					The ``vertices'' and ``contact'' columns evaluate grasps against the closest ground-truth one. 
	}
	\label{tab:grabnet_evaluation}
	\scriptsize	
	\setlength{\tabcolsep}{1pt}
	\begin{tabular}{c|cc|cc|c|c}
		\color{red}
		& \multicolumn{4}{c|}{AMT}                                            & Vertices & Contact			\\ \hline
		& \multicolumn{2}{c|}{Generation} & \multicolumn{2}{c|}{Ground Truth} & mean(cm) & \%      			\\ \hline
		Test Object & mean           & std            & mean            & std             & N=100    & N=20   \\ \hline
		binoculars  & 4.09           & 0.93           & 4.27            & 0.80            & 2.56     & 4.00   \\ \hline
		camera      & 4.40           & 0.79           & 4.34            & 0.76            & 2.90     & 3.75   \\ \hline
		frying pan  & 3.19           & 1.30           & 4.49            & 0.67            & 3.58     & 4.16   \\ \hline
		mug         & 4.13           & 1.00           & 4.36            & 0.78            & 1.96     & 3.25   \\ \hline
		toothpaste  & 4.56           & 0.67           & 4.42            & 0.77            & 1.78     & 5.39   \\ \hline
		wineglass   & 4.32           & 0.88           & 4.43            & 0.79            & 1.92     & 4.56   \\ \hline
		Average     & 4.12           & 1.04           & 4.38            & 0.77            & 2.45     & 4.18   
	\end{tabular}
	\end{center}
\end{table}

\begin{figure}[t]
	\centering
	\includegraphics[trim=000mm 000mm 000mm 000mm, clip=true, width=0.95 \linewidth]{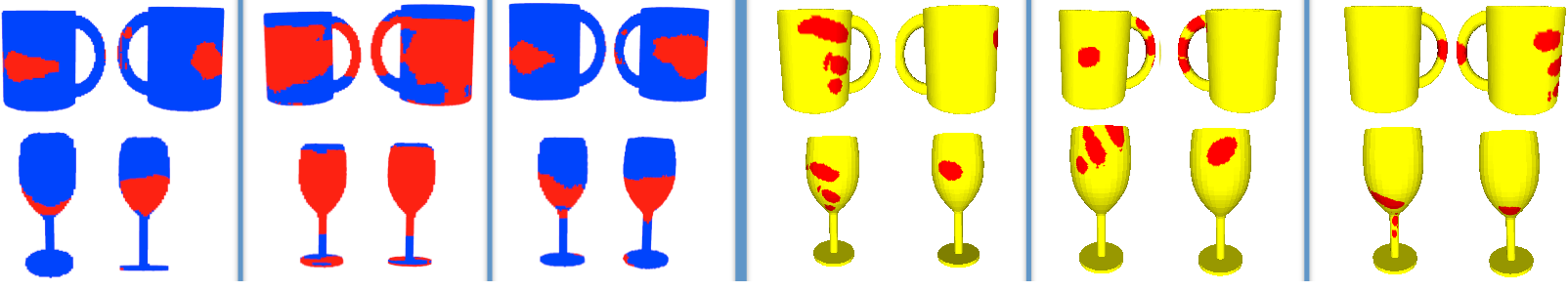}
	\caption{
						Comparison of \grabnet (right) to ContactDB \cite{contactDB_2019} (left). 
						For each estimation we render two views, following the presentation style of \cite{contactDB_2019}. 
	}
	\label{fig:contactdb_comparision}
\end{figure}

\textbf{Evaluation - Contact.}
\highlightFIGTAB{Figure \ref{fig:contactdb_comparision}} shows examples of contact areas (red) generated by \cite{contactDB_2019} \highlightFIGTAB{(left)} and our approach \highlightFIGTAB{(right)}. 
The method of \cite{contactDB_2019} gives only $10$ predictions per object, some with zero contact. 
Also, a hand is supposed to touch the whole red area; this is often not anthropomorphically plausible. 
Our contact is a by-product of \mano-based inference and is, by construction, anthropomorphically valid. 
Also, one can draw infinite samples from our learned grasping latent space. 
For further evaluation, we follow a protocol similar to \cite{contactDB_2019} for our data. 
For every unseen test object we generate $20$ grasps, and for each one we find both the closest ground-truth contact map and the closest ground-truth hand vertices, for comparison. 
Table \ref{tab:grabnet_evaluation} (right) reports the average error over {all} $20$ predictions, in $\%$ for the former and cm for the latter case.

\section{Discussion}

We provide a new dataset to the community that goes beyond previous motion capture or grasping datasets.
We believe that \grab will be useful for a wide range of problems. 
Here we show that it provides enough data and variability to train a novel network to predict {\color{black} human grasping of objects}, as we demonstrate with \grabnet.
But there is much more that can be done.
Importantly, \grab includes the \mbox{whole-body} motion, enabling a much richer modeling than \grabnet. 

\textbf{Limitations:}
By focusing on accurate \mocap, we do not have synced image data. 
However, \grab can support image-based inference \cite{ExPose:2020,Corona_2020_CVPR,kanazawa_cvpr_2018} by enabling 
rendering of synthetic human-object interaction \cite{hasson_2019_obman,multihumanflow,varol17} or 
learning priors to regularize ill-posed inference of human-object interaction from \twoD images \cite{iMapper2018}. 

\textbf{Future Work:}
\grab can support 
learning human-object interaction models 				\cite{iMapper2018,savva2016pigraphs}, 
robotic grasping from imitation							\cite{garciahern2020physicsbased,welschehold2016learning}, 
mapping \mocap markers to meshes 						\cite{oculus2018handMarkers}, 
rendering synthetic images 								\cite{hasson_2019_obman,multihumanflow,varol17}, 
inferring object shape/pose from interaction 			\cite{faisal_HapticSLAM,Oberweger2019}, 
or analysis of temporal patterns 						\cite{pirk2017understanding}.

{
\smallskip
\noindent
\footnotesize
\textbf{\emph{Acknowledgements:}}
We thank S. Polikovsky, M. H\"{o}schle (MH) and M. Landry (ML) for the \mocap facility. 
We thank F. Mattioni, D. Hieber, and A. Valis for \mocap cleaning. 
We thank ML and T. Alexiadis for trial coordination, MH and F. Grimminger for \threeD printing, V. Callaghan for voice recordings and J. Tesch for renderings. 
\textbf{Disclosure:} In the last five years, MJB has received research gift funds from Intel, Nvidia, Facebook, and Amazon. 
He is a co-founder and investor in Meshcapade GmbH, which commercializes 3D body shape technology. 
While MJB is a part-time employee of Amazon, his research was performed solely at, and funded solely by, MPI.
}

\bibliographystyle{splncs04}
\bibliography{bib_GRAB_Paper__Articles,bib_GRAB_Paper__Proceedings,bib_GRAB_Paper__Misc}



\section*{Supplementary material}

# GRAB: A Dataset of Whole-Body Human Grasping of Objects
## *Supplemental Material*

Omid Taheri, Nima Ghorbani, Michael J. Black, and Dimitrios Tzionas

Max Planck Institute for Intelligent Systems, Tübingen, Germany
{otaheri,nghorbani,black,dtzionas}@tuebingen.mpg.de

https://grab.is.tue.mpg.de

## S.1 Supplementary Video

The video on our website presents:

- a narrated overview of our method,
- a wide variety of GRAB sequences (3D moving meshes),
- GrabNet predictions for unseen objects from several viewpoints and
- GrabNet failure cases.

## S.2 GRAB Dataset Content

Our GRAB dataset is available for research purposes on our website. The website contains (at least) the material listed below:

- Our modified version of the object meshes of [2].
- Our marker locations on each object mesh.
- Body shape templates for our subjects.
- Pose parameters for our subjects and objects.
- Code to reproduce the interacting meshes, as seen in our video.
- Per-vertex contact annotations on meshes (body and object) for each frame.
- Vicon MoCap files (labeled marker positions, incl. on the floor and table).

## S.3 Why MoCap Instead of 3D Scan Sequences?

For accurate human shapes, we capture a dense 3D scan for each subject to which we fit a personalized 3D SMPL-X template mesh. However, 3D scanning does not scale up for capturing human-object interaction sequences. This would produce huge amounts of data, the processing of which would be a major undertaking. Moreover, object tracking under occlusions would be still very challenging, as finding scan-to-model correspondences is a hard ill-posed problem. Instead, with MoCap a minimum of 3 marker observations is enough for reliable object pose estimation. The placement of many small markers on the objects means that we

can always estimate object pose. Using MoSh++ for the body, given a ground truth body shape, produces accurate meshes that are on par with 3D scanning but much more practical to capture. We follow therefore this practical and scalable approach; we use a high-end optical MoCap system (Sec. 3.1) and fit full 3D meshes to MoCap markers (Sec. 3.2) for both the human and the object.

## S.4   Why not Capture RGB Images?

Capturing *accurate* human-object interactions while also capturing *natural* RGB images is very challenging. Some recent datasets [4,5] capture hand-only interactions with objects and include RGB images, but the images capture only the hand and not the whole body [4,5] and are not fully natural due to visible instrumentation on the hand [4]. Please note that this latter point is fundamental. Currently one must choose between accurate grasping, which requires instrumentation, or natural images, which reduces the accuracy of ground truth.

Both methods [4,5] suffer from severe hand-object inter-penetrations. Garcia-Hernando et al. [4] originally reconstruct a hand skeleton interacting with 4 object meshes, and their method was reported to have an average *skeleton* penetration depth of $11.0 \pm 8.9$ mm (see Sec. 5.2 of [6]). Similarly, we compute the *surface* penetration between MANO and the 3D object meshes for [5] and find the mean to be $4.36 \pm 0.94$ mm. Although the hand inter-penetration of [5] is not as severe as [4], it suffers from not having realistic contact with objects. In Fig. S.1 we compare the contact "heatmaps" for [5] (left) and GRAB (right). Note that for [5], the thumb and all fingertips are rarely in contact, whereas for GRAB they are frequently in contact. The latter is much more realistic given the central role of the thumb and fingertips in object manipulation. This points to an important technical problem, without lowering the value of these works, as they focus on a challenging application.

We conclude that state-of-the-art interaction methods, that also capture RGB images, suffer from intense occlusions and penetrations along with non-realistic contact between the hand and the objects. Such data is not good to learn an accurate data-driven model of 3D interactions. In contrast, our "use" grasps have only $3.25 \pm 0.68$ mm average *surface* penetration, which is significantly lower than [4,5], and realistic contact between the body and objects, while containing more challenging scenarios, namely dexterous in-hand manipulation and capture of the whole body instead of only the hand.

This is attributed to our precision-focused setup, that increases accuracy on the expense of not capturing RGB images, due to the uniform and artificial texture of the MoCap body suit and the 3D printed objects. We believe that this is a sensible trade-off; one can use our accurate 3D mesh reconstructions to learn a model of 3D interactions, and use it as a prior in future work to improve methods like [4,5] for the hand or the whole body.

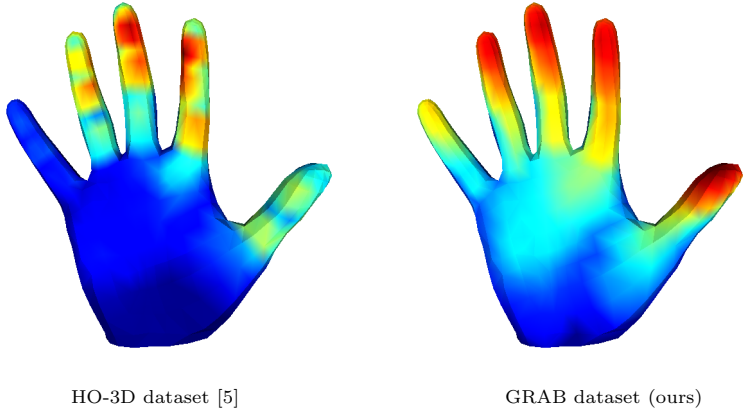

| HO-3D dataset [5] | GRAB dataset (ours) |

**Fig. S.1:** Contact "heatmaps" for HO-3D [5] (left) and GRAB (right), for the right hand. The hotter the color, the more frequently that hand part is in contact with objects. During grasping and manipulation, the thumb finger and all fingertips play a central role. This is evident with GRAB but not with [5].

## S.5 Penetration Plots

In the main paper we describe the observed human-object mesh inter-penetration. Due to space constraints, we report here the corresponding plots in Fig. S.2. We evaluate the degree of penetration for "use" sequences, that pose the most realistic occlusions and capture challenges. "Use" grasps have $3.25 \pm 0.68$ mm average penetration, which effectively corresponds to the missing soft-tissue deformation. Please note that there is no model of the human body with articulated fingers and face that captures such soft-tissue deformation with contact. In addition, 67% of "use" grasps have $\leq 3.5$ mm penetration, 86% $\leq 4.0$ mm, 96% $\leq 4.5$ mm and 99.9% has $\leq 5.8$ mm.

## S.6 Protocol Details

Here we provide details that were not crucial for the main manuscript (Sec 3.4). We capture motions with 4 intents: "use", "pass", "lift", and "off-hand pass". For each sequence we randomize the object position and pose on a resting table, the height of which is also randomized between 75 cm and 120 cm to increase motion variance. We capture the following intents:

**"Use":** For the objects that have a clear everyday use (e.g. drinking from cup), we ask the subject to naturally use them. In case of multiple use cases (e.g. digital/analog photo camera) we capture multiple sequences. For objects without a clear use (e.g. cylinder) the subject has to grasp them and inspect them.

**"Pass":** The subject is asked to pass the object to a predefined direction, that is randomized (e.g. bottom-left, top-right, etc), to increase motion variance.

**"Lift":** The subject is asked to grasp the object, lift it stably in any natural way

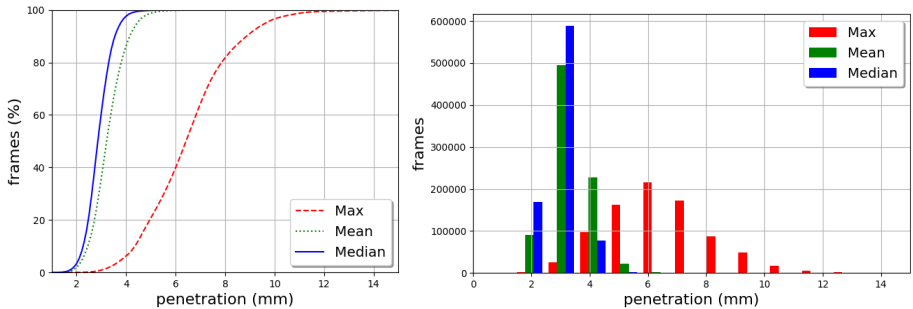

**Fig. S.2:** Penetration plots for "use" grasps. For each frame we store the max (red), mean (green) and median (blue) vertex penetration. **(Left)**: percentage of frames (Y axis) below a varying penetration error (X axis). **(Right)**: bar-plot for number of frames (Y axis) with a specific (quantized for binning) penetration (X axis). The mean penetration is $3.25 \pm 0.68$ mm.

they can imagine, then leave it on the table in any random pose, and repeated this several times with re-grasping. This increases grasp variance, by encouraging the exploration of contact configurations and relative hand-object orientations.
**"Off-Hand pass":** As a form of bimanual manipulation, the subject grasps the object with the off hand, passes it to the dominant hand, and uses it (see "use").

We capture MoCap markers placed on the body, face and fingers, as well as on the object (Sec 3.1 in paper). Additionally, we capture markers attached to the floor and the table, for potential future use. All subjects gave informed consent to share their motion data for research purposes.

## S.7   Computing Contact

Here we provide some additional intuition to Sec. 3.3 and Fig. 4 (right). In particular we explain how we deal with noise in the reconstructed moving meshes to produce clean contact data.

**Figure S.3 (left)**: Consider the illustrated example of a 3D cup. Its mesh thickness is thin, i.e. it has an outer and inner surface that are different, yet close to each other. In Fig. S.3 (top-left) the thumb and index fingers of a grasping hand penetrate both the outer and inner surface. This is due to noise, fitting errors, and because existing models do not model contact-dependent skin deformations.

For these examples the actual contact area is the one on the outer object surface. To annotate only this, following Sec. 3.3, we first compute all colliding triangles and cluster them in connected "rings" (Fig. S.3, top-left). For each "ring" we compute the corresponding penetrating hand areas (Fig. S.3, bottom). The hand areas that contact the inner surface are a subset of the ones that contact the outer one. Then, we remove (big red circles in Fig. S.3 bottom) the

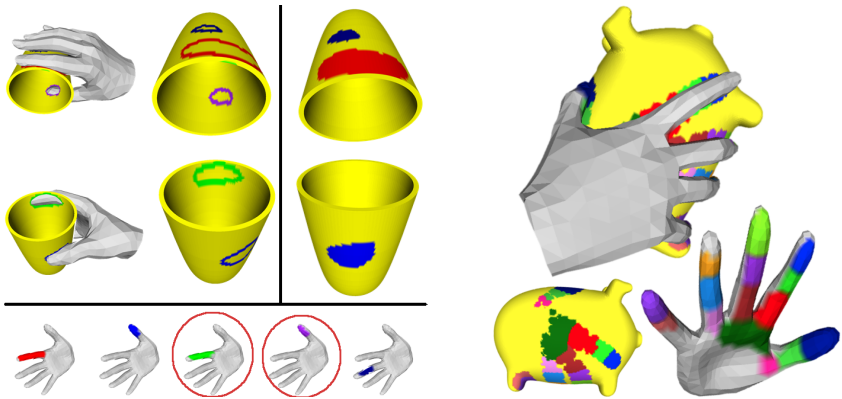

**Fig. S.3: (Left)**: Annotating contact areas for a hand grasping a cup: (top-left) "rings" of colliding triangles, color-coded for each finger, (bottom) penetrating hand areas that correspond to each "ring"; the ones corresponding to the green and purple vertices (red circled hand parts) penetrate the inner cup surface and are ignored, (top-right) the final filtered "rings" and the enclosed vertices are annotated as contact areas. The contact labels are binary; color is used here only for visualization purposes. **(Right)**: Contact labels can though be more fine-grained, e.g. using the contacting hand parts or hand vertices. Here we see an example of the former case. (top) Each color represents a contact area caused by a different hand part. (bottom) Contact areas are shown also on the object and unposed hand for clarity. Note that the size of contact areas is expanded for illustration purposes.

purple and green groups, we keep only the remaining "rings", and annotate the vertices enclosed by them as contact vertices (Fig. S.3, top-right).

**Figure S.3 (right)**: The above procedure gives binary contact annotations ("contact" on "not in contact"). Contact labels, however, can be more fine-grained, e.g. with the label of the corresponding hand part (Fig. S.3, right), or even with the point on the 3D hand surface (Fig. S.4). For the former example, we find the object vertices that are in contact, and for each one we find the closest SMPL-X/MANO bone, and assign its ID as the contact label.

## S.8  Adapting MoSh++

We adapt MoSh++ [9] (Sec. 3.2) for capturing the whole body (including the hands and face). The human and object are tracked independently and on a per-frame basis, for simplicity. We make two small changes to MoSh++. First we use the ground-truth body shape, obtained from a 3D scan. Consequently, we do not use MoSh++ to estimate body shape. Second, we extend MoSh++ to estimate the parameters of the SMPL-X body model. This means extending it to capture facial pose and expression parameters. Additionally, we estimate the rigid 6 DoF pose of the objects using their known shape and the detected markers.

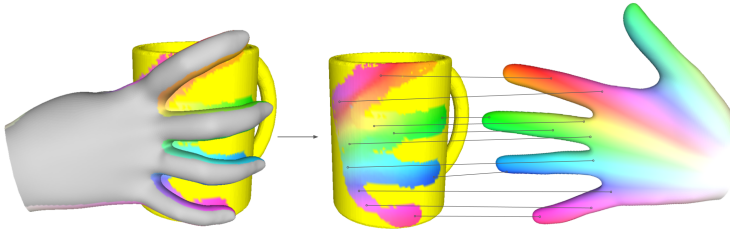

**Fig. S.4: (Right)**: Fine-grained contact labels. In contrast to the binary contact labels of Fig. 4 (left) and Fig. 9 of the main manuscript, and the part-based contact labels of Fig. S.3, here we show an example of much more fine-grained labels. **(left)** A 3D hand-object grasp configuration. **(middle)** The object alone. **(right)** The hand in canonical pose. We highlight different points on the 3D surface of the inner hand with color gradients. The contact between the hand and object define surface correspondences between them (shown as lines).

To adapt MoSh++ to capture faces, we need to tune the parameters of the model. For this [9] follows a data-driven approach; they capture the SSM dataset with an optical MoCap system synchronized with a 3D body scanner, and use the scans for computing a reconstruction quality metric. However, SSM has markers only on the main body, while also the fingers of the scans are very noisy.

Capturing such a dataset, with clear scan regions for both the body, the face and all fingers, as well as synced MoCap for them, is too challenging. Instead, we follow a more practical approach, and create a synthetic dataset by animating SMPL-X and generating virtual markers on the moving meshes. To bridge the domain gap, we simulate noise for marker position and visibility; we randomly add 3D Gaussian noise with 1 mm variance in marker positions, as in [9], and randomly drop up to 5 markers per frame.

Unfortunately, there is no existing dataset with rich SMPL-X sequences. However, its model formulation is compatible to SMPL [8] for the body, FLAME [7] for the face, and MANO [13] for the hands. Therefore, we resort to datasets specific to each part to animate the body, face and hands. For the *body*, we employ DFAUST [1] that captures 10 subjects performing 10 sequences each. We split the subjects into 6 for training and 4 for a withheld test set. We compute personalized SMPL-X mesh templates by registering the model to one scan per person as in [13, 10], and pose their body according to the registrations of DFAUST. For the *hand*, we employ the hand-only MANO model registrations of [13]. From the 1554 hand poses we hold out 155 for the test set and use the rest for training. We then add hand motion to each body sequence by randomly choosing 15 hand poses and interpolating between them. For the *face*, we employ sequences of FLAME parameters from [3, 12]; the latter covers extreme facial expressions, while the former has everyday speaking expressions. We randomly choose 100 sequences from each dataset, splitting them into 60 for the training set and 40 for the withheld test set.

**Table S.1:** Evaluation of MoSh++ on the synthetic dataset. We compare the vanilla [9] to our adapted version. For the first stage of MoSh++, Stage-I, we report the distance of the latent marker placement compared to ground-truth marker locations, and for the second stage of MoSh++, Stage-II, we report the average vertex-to-vertex error between estimated and ground-truth meshes.

| Mosh++ version | MoSh Stage-I | | MoSh Stage-II | | |
|---|---|---|---|---|---|
| | mean ± std | median | mean ± std | median | |
| Vanila [9] | 4.76 ± 1.03 | 4.55 | 5.59 ± 1.86 | 5.28 | mm |
| **Our adapted** | **3.09 ± 0.55** | **2.80** | **4.86 ± 1.83** | **4.48** | |

We use this dataset to set the weights following the approach in [9]. Table S.1 compares a standard version of [9] with our adapted version on the synthetic test set for both stages of MoSh++. For the first stage of MoSh++ (Stage-I) we report the distance of the latent marker placement compared to ground-truth marker locations in mm. For this stage we start from random marker placement guesses in the 1-ring neighborhood of the ground-truth locations. We repeat this three times with different random seeds for selecting 12 frames of MoSh++; see [9]. For the second stage (Stage-II), we use the optimized latent marker placements resulting from each random seed of the Stage-I and report the average vertex-to-vertex error between estimated and ground-truth meshes in mm. In each stage we choose the wights that minimizes the reported error. Our adapted version shows a clear improvement, by fitting the whole body, hands and face, with weights $\lambda$ tuned on our synthetic dataset. In contrast [9] tunes only the body weights on their SSM dataset, it fits the hands with empirical weights, and does not fit the face.

## S.9 GrabNet

To show that GRAB can be useful for machine learning applications, we train on it a generative model to generate grasping poses for a 3D object, that we call GrabNet. For this example application we focus only on right-hand grasps for simplicity, but GRAB provides much richer data than this.

### S.9.1 Data Preparation

We use only right hand data due to the large size of GRAB, but left hand poses could also be mirrored to appear as right ones for data augmentation. In order to select right-hand frames for training GrabNet we use the following rules. (i) The right hand should be in contact. (ii) The left hand should not have any contact. (iii) The object's vertical position should be at least 5 mm different from its initial one (i.e. it should be lifted from the resting table). (iv) The right thumb and at least one more finger should be in contact. (v) A finger is considered a contacting finger, when it is in contact with at least 50 object vertices. With

these filters we make sure that we have only stable grasps with which to train GrabNet.

To model arbitrary shapes, we use the basis point set [11] representation $BPS_o$ for all our objects. For computational efficiency, we precompute $BPS_o$ and load it from memory during training. We sample basis points in a sphere of 150 mm radius, that is big enough to cover our centroid-centered objects. We empirically found 4096 basis points to be enough. We then compute the distances from the basis points to our object meshes.

Out of our 51 objects, borrowed from [2], we hold out 4 objects for the validation set ("apple", "toothbrush", "elephant" and "hand"), 6 objects for the test set ("mug", "wineglass", "camera", "binoculars", "frying pan" and "toothpaste"), and use the remaining 41 objects for the training set.

The training, validation and test splits contain roughly 320k, 31k and 65k data points, correspondingly.

To prepare the training data for RefineNet we add Gaussian noise to the Ground Truth MANO parameters of the selected data for GrabNet. Since the perturbation need to be minimal we empirically find $\mathcal{N}(\mu = 0, \sigma^2 = 0.2)$, $\mathcal{N}(\mu = 0, \sigma^2 = 0.004)$, and $\mathcal{N}(\mu = 0, \sigma^2 = 0.05)$ for MANO finger joints rotation, root rotation, and translation respectively.

### S.9.2 Results: Success and Failure Cases

Figures S.6, S.7 and S.8 provide a wide variety of qualitative GrabNet results. More specifically, they show 10 different grasps (rows) generated for 6 unseen objects (columns). The three figures show three different viewpoints (one view per figure) for the same grasp of the 10 × 6 grid. We see that most grasps look natural and plausible, as GrabNet is learned from high-quality GRAB captures. More results with a rotating viewpoint are shown in the video on our website.

GrabNet can still generate some failure cases. These are mostly cases of penetrating fingers; there are not many cases of contacting fingers that fly away from the object. Penetrations are observed mostly for objects with thin parts (cup handle, wine glass, bowl). We found the frying pan to be the most challenging object, due to its comparably big size along with its thin surface walls and handle. This might be due to the sparse $BPS_o$ representation for 3D object shapes capturing mostly their bigger parts. Furthermore, at the moment we use a penetration and a contact term in the training loss of GrabNet as soft constraints, since here we focus on a data-driven method. One could add an optimization stage to refine the regressed grasp with hard penetration and contact constraints.

The results show the value of GRAB for training data-driven models, but also point to room for improvement for GrabNet's modeling and training scheme.

### S.9.3 GrabNet Implementation Details

The architecture for GrabNet is shown in Fig. S.5. For CoarseNet, we concatenate the object $BPS_o$ representation with MANO hand parameters as input to the

encoder, and also concatenate it with the latent code as the input (condition) to the decoder. The outputs of the decoder are MANO translation ($\gamma \in R^3$) and joint angles ($\theta \in R^{96}$) in the continuous 6-dimensional representation of [15].

Using our validation set, we found out that 16 dimensions for the latent space results in generating better grasps. Qualitative results are provided in Fig S.6.

For RefineNet we take the output of the CoarseNet (MANO parameters) and first compute the distances of MANO vertices to the object vertices. We then pass the distances with the MANO parameters to the network. RefineNet refines the input grasp through 3 iterations. The CoarseNet and RefineNet are trained for 16 and 23 epochs respectively with the learning rate starting from $5e-4$, decreasing on validation error plateau to 0.1 times, and early stopping after 8 epochs with no improvement in validation error. Both networks are trained separately.

### S.9.4 Filtering out Unreliable Turkers

As mentioned in the main paper (Sec. 4), along with ground-truth (GRAB) and GrabNet-generated grasps, we pass to Turkers noisy grasps generated by perturbing ground-truth ones. These noisy grasps are our test for spotting unreliable Turkers, that either select their answers randomly or misunderstand the task. Specifically, we remove the ones that gave a rating of 3 or more (indicating good realism) for at least 20% of these noisy grasps. In total we removed 54 out of 170 Turkers.

### S.9.5 Heatmaps for Various Intents and Fine-Grained Numbers

Similar to Sec 3.5 of the main manuscript, here Fig. S.9 provides additional fine-grained numbers for in-contact parts of the body. Each row corresponds to an intent in the GRAB dataset. For each intent, the right column shows the contact percentage and "heatmap" for the right hand, left hand, and head across all frames and relative to *all* body vertices. In the three left columns, the "heatmaps" and percentages are relative to only *each part's* vertices and for only the frames for which these parts are in contact (left hand, right hand, and head), for visualization purposes. For example, for the "use" sequences (second row in Fig S.9), the right hand was in contact for 90.62% of all frames, and in those contact frames the thumb fingertip was in contact for 99.88% of them.

## S.10 Bias from MoCap Markers

A natural question arises - are subjects biased in their grasps by MoCap markers? We empirically place more markers in areas less likely to be contacted, according to object affordances. To account for potential occlusions, though, we have to place some markers in other areas as well. For this reason, we still expect our markers to be contacted.

Our subjects did not complain about discomfort or bias, yet we need more evidence for this. Apart from the analysis in the main manuscript (see Sec 3.5 and Fig. 5), here we perform k-means clustering (k=20) on our grasps, and visualize each cluster center, i.e. a grasping hand, and the grasped object. We observe that several clusters (typically 3-6 out of 20) show that fingers do come in contact with markers. Figure S.10 shows for 5 objects (rows) 3 representative contacting clusters (columns). We believe this is good additional empirical evidence that our 1.5 mm radius hemi-spherical markers cause no or minimal bias.

### S.11    Influence of Contact Heuristic Thresholds

We use several heuristics to determine contact frames, see Sec. 3.3 in the main manuscript. For the contact "heatmap" analysis we take all the contact frames for which object is being manipulated, i.e. it is off the table. Because the heatmap is integrated over many frames, small variations in the heuristics have little impact on the contact patterns.

To show this empirically, we perform a sensitivity analysis by changing our thresholds. Figure S.11 shows "heatmaps" for "use" sequences for several setups, following the format of Fig. 5 of the main manuscript. The results verify our hypothesis that the heuristics have minimal influence.

### S.12    Influence of Object Size

Figure S.12 shows the effect of object size on hand contact for 5 shape primitive objects. Smaller objects are grasped mainly by the dominant hand, while, for larger objects, the other hand helps too. Small-sized subjects have a large tendency for bimanual grasps, and involve more finger and contact around metacarpophalangeal joints.

### S.13    Vicon Software

We use the Vicon system with the Shōgun [14] software, in two main modes. Shōgun-Live is used for camera calibration and synchronization, as well as for subject calibration and data capture. The output consists of 3D marker locations and labels (IDs). However, both can be noisy; markers can disappear due to extreme occlusions, ghost markers might appear due to reflections, while labels can be mistakenly swapped when many markers come close together. This calls for the need for data cleaning in a post-processing stage. This is important for strong interactions, as getting even only 3 cleaned marker observations is enough to solve for the rigid pose of a manipulated object. Shōgun-Post is used for such semi-automatic data cleaning by 4 trained annotators.

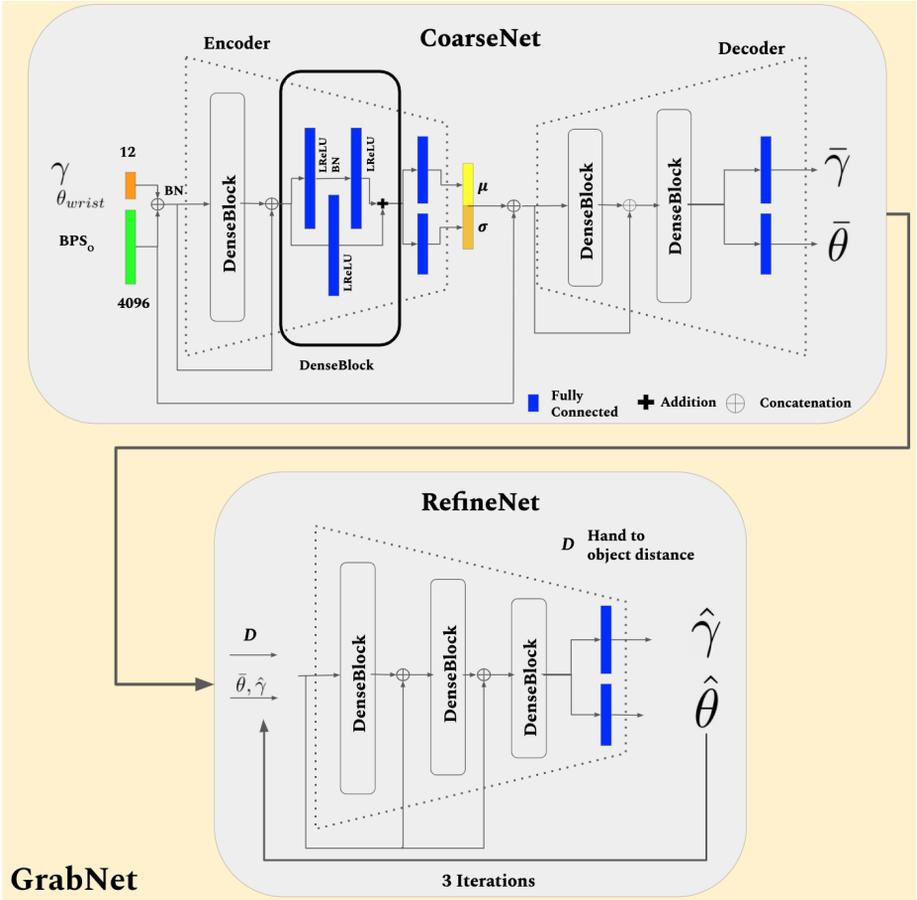

**Fig. S.5:** GrabNet Architecture. For the encoder input, we concatenate the BPS representation of the object with MANO parameters, while for decoder input we concatenate it with a sample from latent space. The decoder gives the MANO hand parameters which we pass to the MANO model to obtain the 3D hand mesh.

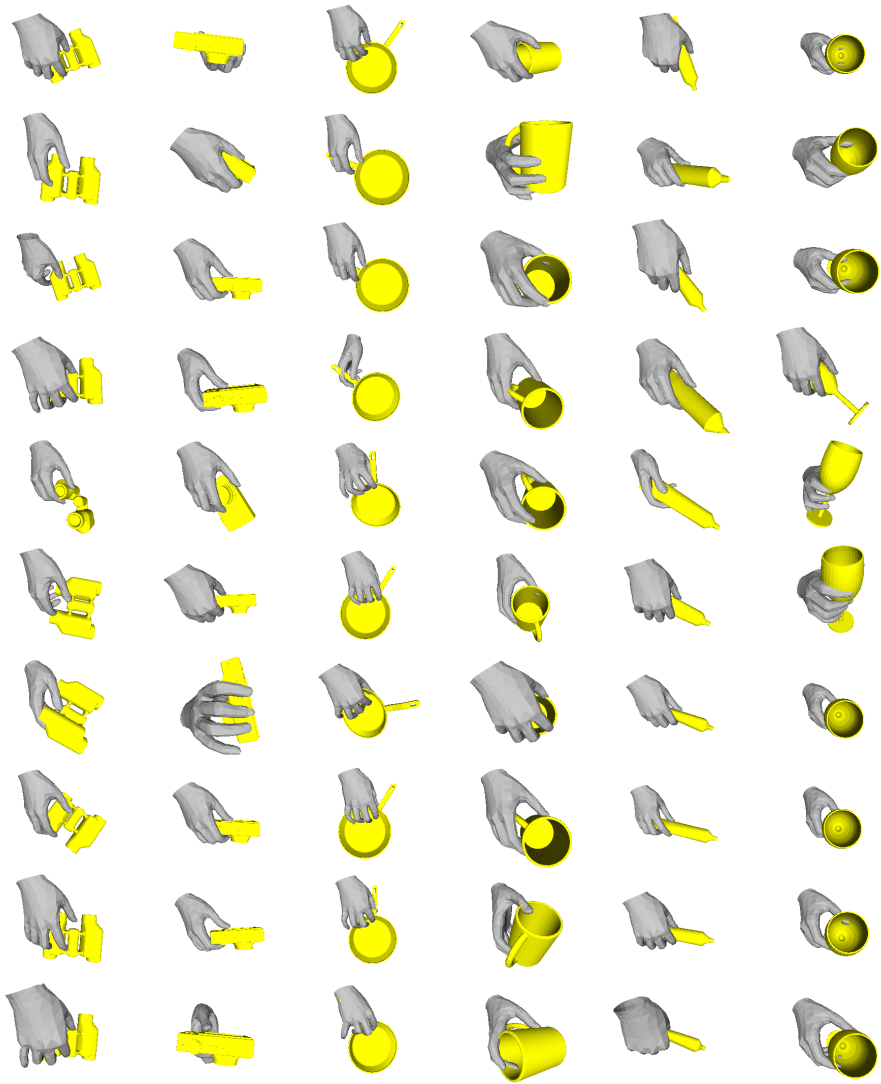

**Fig. S.6:** Visualization of 10 different grasps (rows) generated by GrabNet for 6 unseen objects (columns). Conditioned on the $BPS_o$ representation of unseen 3D object shapes, we sample from the learned 16 dimensional latent grasping space $Z$ of CoarseNet (see Fig. 8 of the paper). We then concatenate $BPS_o$ to the $Z$ sample, and pass them to the decoder of CoarseNet, that outputs the coarse grasping MANO hand model parameters. We then pass the coarse grasps to the RefineNet to get the final grasps. Some failure cases are highlighted with red. Different viewpoint for the results of Fig. S.7, S.8.

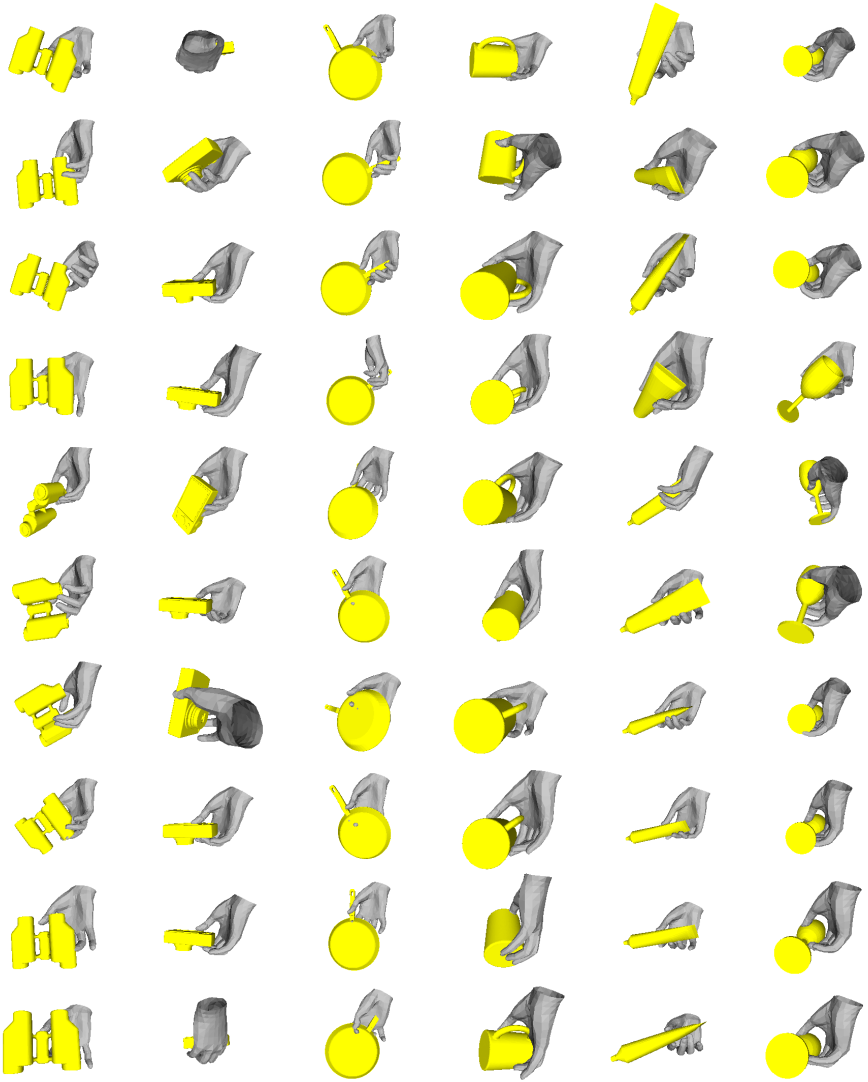

**Fig. S.7:** Visualization of 10 different grasps (rows) generated by GrabNet for 6 unseen objects (columns). Conditioned on the $BPS_o$ representation of unseen 3D object shapes, we sample from the learned 16 dimensional latent grasping space $Z$ of CoarseNet (see Fig. 8 of the paper). We then concatenate $BPS_o$ to the $Z$ sample, and pass them to the decoder of CoarseNet, that outputs the coarse grasping MANO hand model parameters. We then pass the coarse grasps to the RefineNet to get the final grasps. Some failure cases are highlighted with red. Different viewpoint for the results of Fig. S.6, S.8.

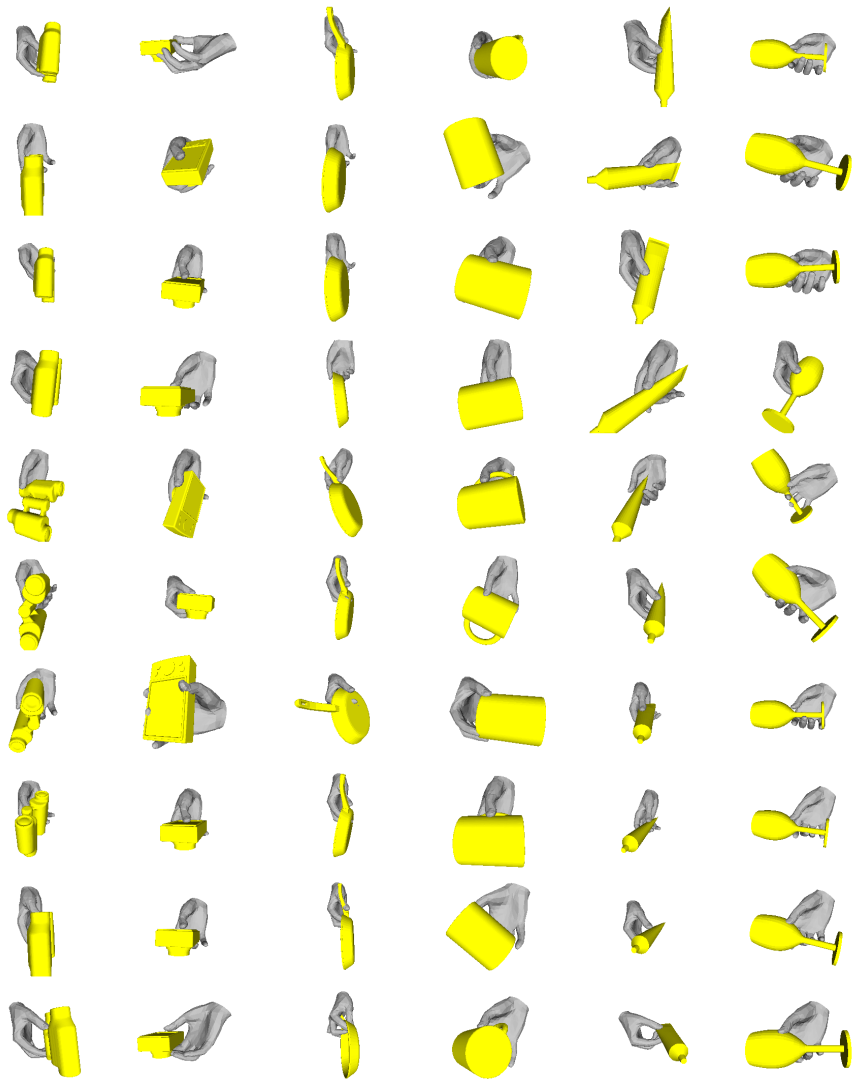

**Fig. S.8:** Visualization of 10 different grasps (rows) generated by GrabNet for 6 unseen objects (columns). Conditioned on the $BPS_o$ representation of unseen 3D object shapes, we sample from the learned 16 dimensional latent grasping space $Z$ of CoarseNet (see Fig. 8 of the paper). We then concatenate $BPS_o$ to the $Z$ sample, and pass them to the decoder of CoarseNet, that outputs the coarse grasping MANO hand model parameters. We then pass the coarse grasps to the RefineNet to get the final grasps. Some failure cases are highlighted with red. Different viewpoint for the results of Fig. S.6, S.7.

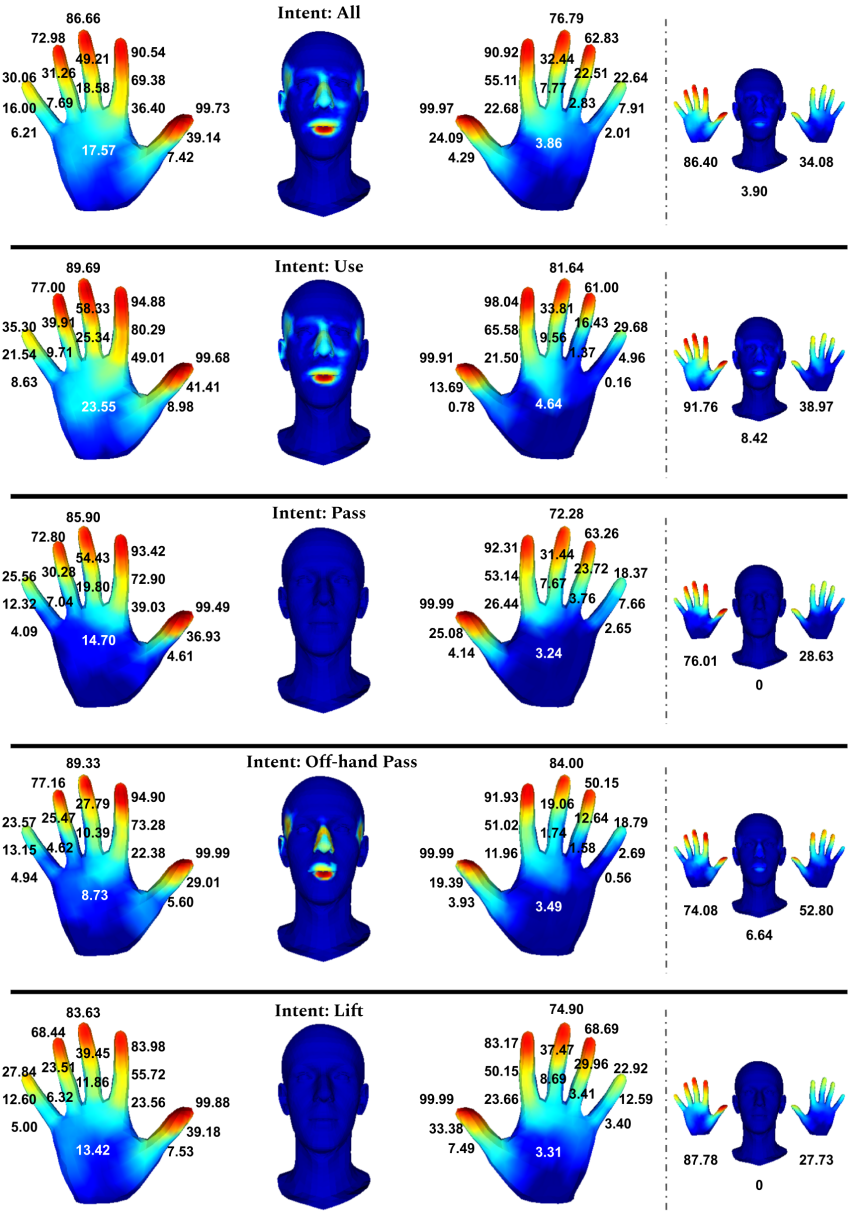

**Fig. S.9:** Contact "heatmaps" and percentages for all intents in GRAB, for various body parts. Each row corresponds to an intent in the GRAB dataset. For each intent, the right column shows the results for each part (right hand, left hand, and head) across all frames and relative to *all* body vertices. In the three left columns, the results are relative to only *each part's* vertices and for only frames for which these parts are in contact (left hand, right hand, and head), for visualization purposes.

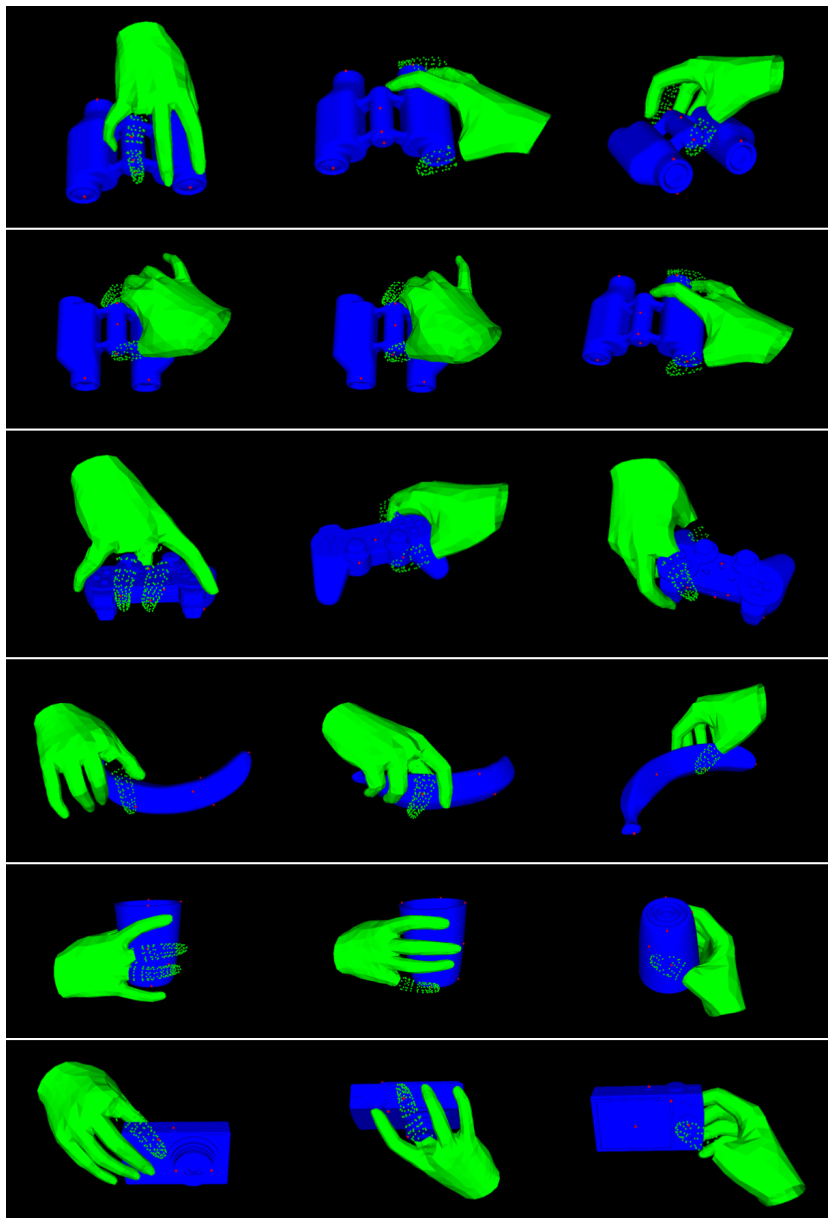

**Fig. S.10:** Do subjects avoid markers? To answer this, we perform k-means clustering (k=20) on our grasps, and visualize each cluster center, i.e. a grasping MANO (green), and the grasped object (blue). We observe that several cluster centers (columns) per object (rows) show that subjects contact MoCap markers (red); here we show 3 clusters for 5 objects. For fingers that contact markers we render only the vertices, to allow to see the markers (best viewed on screen).

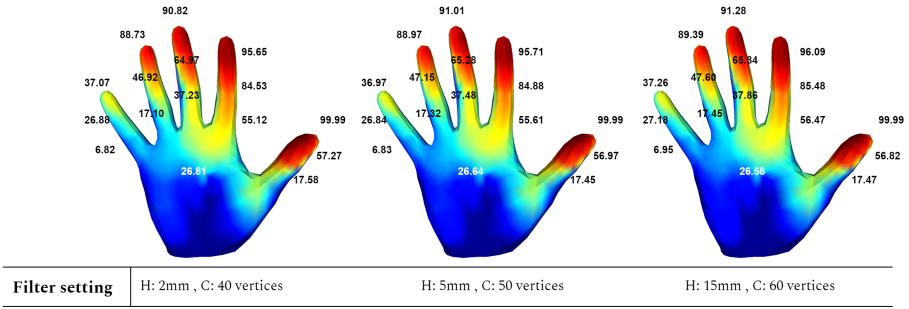

| Filter setting | H: 2mm , C: 40 vertices | H: 5mm , C: 50 vertices | H: 15mm , C: 60 vertices |

**Fig. S.11:** Sensitivity analysis for contact heuristics. We follow the format of Fig. 5 of the main manuscript, and show "heatmaps" and contact likelihoods in percentages % for a subset of "all" sequences with different setups (columns), as indicated in the labels. The symbol $H$ denotes the minimum difference between the object's vertical position from its initial one (resting on a table). The symbol $C$ denotes the minimum number of object vertices that we require to be in contact with each finger. The figure shows that threshold choices have minimal effect when integrated over many frames to create "heatmaps".

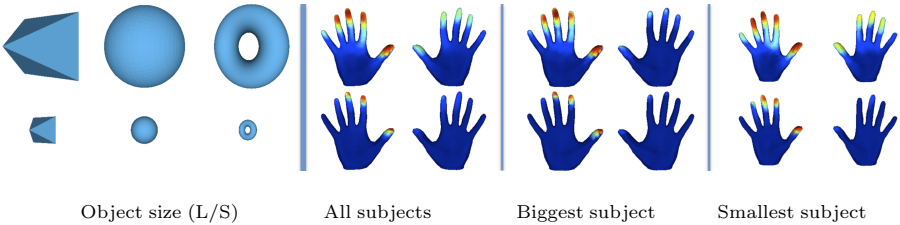

Object size (L/S)  All subjects  Biggest subject  Smallest subject

**Fig. S.12:** Effect of object size on contact during grasping. We show contact "heatmaps" for all subjects, the biggest subject, and the smallest subject.



\end{document}